\documentclass[10pt,twocolumn,letterpaper]{article}

\usepackage{cvpr}              

\usepackage[dvipsnames]{xcolor}

\definecolor{cvprblue}{rgb}{0.21,0.49,0.74}
\usepackage[pagebackref,breaklinks,colorlinks,citecolor=cvprblue,linktocpage=true]{hyperref}
\usepackage{dsfont}
\usepackage{bm}
\usepackage{amsmath}
\usepackage{multirow}
\usepackage[ruled]{algorithm2e}  
\usepackage{titletoc}
\def\paperID{6869} 
\def\confName{CVPR}
\def\confYear{2024}

\title{Global and Local Prompts Cooperation via Optimal Transport \\
for Federated Learning}

\author{Hongxia Li$^1$\quad  Wei Huang$^2$\quad Jingya Wang$^1$\quad Ye Shi$^{1,}$\thanks{Corresponding author.} \\
$^1$ShanghaiTech University, Shanghai, China\\
$^2$RIKEN Center for Advanced Intelligence Project, Japan\\
{\tt\small \{lihx2,wangjingya,shiye\}@shanghaitech.edu.cn, wei.huang.vr@riken.jp}\\
{\small {\url{https://github.com/HongxiaLee/FedOTP}}}
}

\begin{document}
\maketitle

\begin{abstract}
Prompt learning in pretrained visual-language models has shown remarkable flexibility across various downstream tasks. Leveraging its inherent lightweight nature, recent research attempted to integrate the powerful pretrained models into federated learning frameworks to simultaneously reduce communication costs and promote local training on insufficient data. Despite these efforts, current federated prompt learning methods lack specialized designs to systematically address severe data heterogeneities, e.g., data distribution with both label and feature shifts involved. To address this challenge, we present Federated Prompts Cooperation via Optimal Transport (FedOTP), which introduces efficient collaborative prompt learning strategies to capture diverse category traits on a per-client basis. Specifically, for each client, we learn a global prompt to extract consensus knowledge among clients, and a local prompt to capture client-specific category characteristics. Unbalanced Optimal Transport is then employed to align local visual features with these prompts, striking a balance between global consensus and local personalization. By relaxing one of the equality constraints, FedOTP enables prompts to focus solely on the core regions of image patches. Extensive experiments on datasets with various types of heterogeneities have demonstrated that our FedOTP outperforms the state-of-the-art methods. 

\end{abstract}

\section{Introduction}
\label{sec:intro}
Federated learning \cite{mcmahan2017communication} is a distributed machine learning framework that enables decentralized collaboration among participants without sharing their training data. However, current federated learning methods involve high training and communication costs due to the need to update and share model parameters with the server. This constraint has typically restricted these methods to modest backbone architectures, hindering their feature capacity and resulting in performance limitations and training instability \cite{yang2023efficient}.

Recently, vision-language pre-trained models like Contrastive Language-Image Pretraining (CLIP) \cite{radford2021learning} have shown potential in learning robust and versatile representations suitable for various image distributions, aligning with the objectives of federated learning. However, the substantial communication overhead between the server and clients renders training CLIP in federated learning frameworks. Besides, overfitting concerns may arise when large-scale models are trained with limited client data. Prompt learning \cite{liu2023pre, zhou2022learning} provides a flexible way to adapt pre-trained models to downstream tasks by training only additional parameters. This enables prompts to capture task-specific information while guiding the fixed model's performance. Leveraging its lightweight nature, prior research \cite{zhao2022reduce, guo2023promptfl} has explored the integration of prompt learning into federated learning to overcome the problems outlined above. 

In real-world scenarios, client data often exhibits variations in domain discrepancies (feature shift) \cite{li2020fedbn} or imbalanced class distributions (label shift) \cite{li2021model}. 
Simply applying the FedAvg \cite{mcmahan2017communication} method on prompts \cite{guo2023promptfl} across all clients tends to deviate from their local distribution, leading to unsatisfactory performance. Hence, it's crucial to develop specialized personalized federated prompt learning approaches to effectively address data heterogeneity. 
pFedPrompt \cite{guo2023pfedprompt} introduced personalization into federated prompt learning by maintaining personalized attention modules to generate spatial visual features locally while learning user consensus through shared text prompts. However, in the presence of a significant label shift or notable feature shift, merely learning a shared prompt in the language modality is inadequate. 

To resolve these limitations, we propose simultaneously learning both a shared global prompt and a personalized local prompt for each client in the local training phase. After local training, the local prompt remains on the client side, while the global prompt is transmitted to server to aggregate with prompts from other clients. In this manner, the client owns the capacity to acquire consensus knowledge among clients from the global prompt, while also being able to discern client-specific user traits through the local prompt. 

To further achieve a balance between global consensus and local personalization, we introduce Federated Prompts Cooperation via Optimal Transport (FedOTP). FedOTP utilizes Optimal Transport (OT) \cite{kantorovich2006translocation} to align local visual features with both global and local textual features through an adaptive transport plan, promoting fine-grained matching across vision and language modalities and strengthening collaboration between the global and local prompts. The adaptive OT transport plan can provide resilience to visual misalignment and effective adaptation to feature shifts. 
It's worth noting that the standard OT imposes two hard equality constraints on the transport plan, leading to each image patch being assigned to prompts. This may potentially cause prompts to capture some class-irrelevant information from the image and consequently influence the final results. To mitigate this, we consider employing unbalanced OT by relaxing one of the equality constraints, allowing prompts to focus solely on the most relevant image patches rather than the entire content of the image. For an efficient solution, we apply a fast implementation of Dykstra's algorithm \cite{dykstra1983algorithm} in our FedOTP, enabling swift convergence and focusing on the core area of the image during iterations. 

Our main contributions are summarized as follows: 
\begin{itemize}
    \item We are the first to explore the mechanism of prompts' cooperation in federated learning where severe data heterogeneity is present. More precisely, we train both a global prompt for consensus across clients and a local prompt to capture client-specific category traits simultaneously.
    \item We propose FedOTP, a federated learning framework utilizing unbalanced OT to enhance the cooperation between global and local prompts. Through unbalanced OT, we align local visual features with textual prompts while enable prompts to focus solely on the critical image patches.
    \item We conducted extensive experiments on widely adopted datasets in various data heterogeneity with feature shifs and label shifts, and significant result improvement verifies the superiority of our FedOTP. In addition, we demonstrated the ability of FedOTP to balance consensus and local personalization through visualizations.
\end{itemize}

\section{Related Work}
\label{sec:relatedwork}

\subsection{Personalized Federated Learning} 
Personalized federated learning (PFL) is a highly regarded research field because of its potential to address statistical and systemic heterogeneity across clients. Various approaches have been proposed in prior research to achieve PFL. The most common method involves the inclusion of regularization terms in the loss function \cite{li2020federated, li2021ditto, t2020personalized}, and fine-tuning the global model on clients' local datasets \cite{wang2019federated, mansour2020three, fallah2020personalized, khodak2019adaptive}.
Additionally, some works focus on explicitly seeking a trade-off between the global model and the local models \cite{hanzely2020federated, liang2020think, marfoq2022personalized, chen2021bridging}. To enhance adaptability to diverse data distributions, certain techniques have delved into clustering methods for client grouping \cite{sattler2020clustered, huang2021personalized,vahidian2023efficient}. Leveraging the relationships and data distribution among clients, methods like FedPAC \cite{xu2023personalized}, and FedDisco \cite{ye2023feddisco} introduce novel weighted aggregation techniques to promote intensive collaboration among similar clients. Furthermore, some researchers have explored the decomposition of model parameters into base layers and personalized layers. For instance, FedPer \cite{arivazhagan2019federated}, FedRep \cite{collins2021exploiting}, and FedBABU \cite{oh2021fedbabu} learn personalized classifier heads locally while sharing the base layers, and FedTP \cite{li2023fedtp} learns personalized self-attention layers for each client. The importance of PFL was pointed out theoretically by \cite{huang2023understanding}.

The methods mentioned above primarily target label shift data heterogeneity. However, they may not perform well when substantial domain differences exist among clients. In dealing with these feature shifts, FedBN \cite{li2020fedbn} employs local batch normalization to mitigate the feature shift before model averaging, while PartialFed \cite{sun2021partialfed} extends this strategy by selecting personalized parameters according to distinct feature traits of different clients. Besides, FedPCL \cite{tan2022federated} enhances each client's ability to utilize pre-trained models by extracting client-specific and class-relevant information. Our FedOTP explores the cooperation between global and local prompts to effectively address both label shift and feature shift data heterogeneity. 

\subsection{Prompt-based Federated Learning} 
Prompt learning, originating from NLP models, has expanded to Vision Language Models. Initial methods like CLIP \cite{radford2021learning} involved manually crafted templates, while recent approaches concentrate on learning prompts in a continuous embedding space. CoOp \cite{zhou2022learning} fine-tunes CLIP with continuous prompt vectors. Based on this, plenty of studies \cite{zhou2022conditional, he2022hyperprompt, chen2022prompt, lu2022prompt, khattak2023maple, liu2023hierarchical} have been introduced to enhance the effectiveness of prompt learning.
To accelerate the global aggregation and handle situations with insufficient user data, FedPrompt \cite{zhao2022reduce} and PromptFL \cite{guo2023promptfl} have introduced prompt learning into Federated Learning. Based on these two works, several methods have made substantial progress in various domains. For instance, FedPR \cite{feng2023learning} focuses on learning federated visual prompts within the null space of the global prompt for MRI reconstruction.
Based on CLIP, FedAPT \cite{su2022cross} introduces a federated adaptive prompt tuning algorithm for cross-domain federated image classification, and FedCLIP \cite{lu2023fedclip} utilizes an attention-based adapter to optimize the utilization of pre-trained model information. To tackle statistical heterogeneity among clients, pFedprompt \cite{guo2023pfedprompt} maintains a non-parametric personalized attention module for each client to generate locally personalized spatial visual features, 
and pFedPG \cite{yang2023efficient} designs a client-specific prompt generator at the server to create personalized prompts. 
While these works show the potential of prompt learning in Federated Learning, there remains a deficiency in technical enhancements tailored to PFL scenarios. 
Compared with these methods, our FedOTP employs OT to balance the global consensus and local personalization from the collaboration of global and local prompts.

\begin{figure*}[ht]
\centering
\includegraphics[width=1\textwidth]{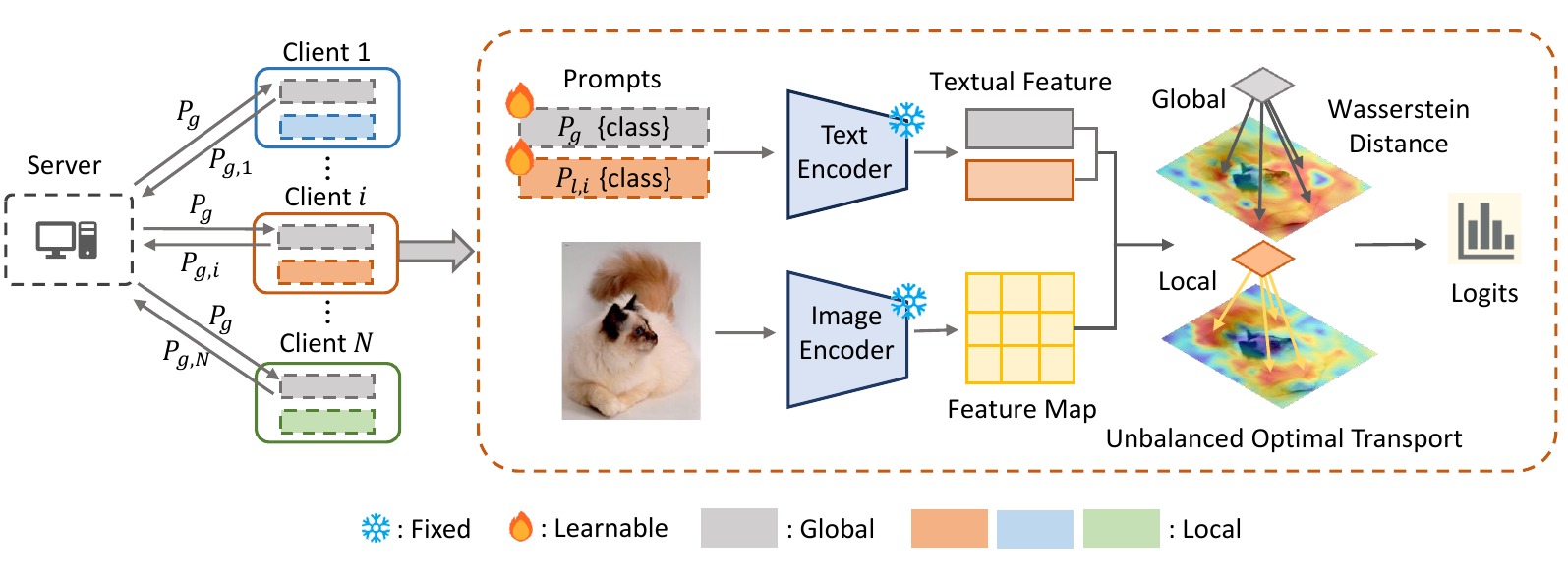} 
\caption{Overview of our FedOTP. On the left, clients transmit global prompts to the server for aggregation while retaining local prompts locally. The right shows the workflow of Global-Local prompt cooperation mechanism, which employs unbalanced Optimal Transport to align visual feature maps with each prompt. }
\label{pipeline}
\end{figure*}

\subsection{Optimal Transport} 
Initially developed as a solution to optimize the cost of moving multiple items concurrently, Optimal Transport (OT) \cite{kantorovich2006translocation} has gained significant attention in the machine learning and computer vision community.
To accelerate the convergence and efficiently deal with large-scale problems, \cite{cuturi2013sinkhorn} introduced Sinkorn's algorithm \cite{sinkhorn1967diagonal} for computing an approximate transport coupling with entropic regularization.
Unbalanced OT relaxes the equality constraint of classical OT, allowing for partial displacement in the transport plan. It can be efficiently solved using the generalized Sinkhorn's algorithm \cite{chizat2018scaling} by incorporating soft penalties based on Kullback-Leibler divergence \cite{frogner2015learning} or parallel algorithm \cite{lahn2024combinatorial}.
Given its remarkable ability in distribution matching, OT has been applied in various theoretical and practical tasks, including domain adaptation \cite{damodaran2018deepjdot, fatras2021unbalanced, chang2022unified, li2024unsupervised}, learning with noisy labels \cite{feng2023ot,chang2023csot}, causal discovery \cite{li2021causal,torous2021optimal,tu2022optimal}, federated learning \cite{farnia2022optimal,chiang2023optimal}, outlier detection \cite{phatak2022computing} and so on. In prompt learning field, PLOT \cite{chen2022prompt} proposes to learn multiple prompt sets for diverse contextual representations and use OT to align the features of vision and language modalities.
Different from PLOT, we employ unbalanced OT to enhance the cooperation between global and local prompts by relaxing one of the equality constraints, which allows prompts to concentrate exclusively on the most relevant image patches.

\section{Preliminaries}
\label{sec:preliminaries}

\subsection{Prompt Learning}
To adapt pre-trained models like CLIP \cite{radford2021learning} to downstream tasks, prompt learning methods \cite{zhou2022learning, zhou2022conditional} provide an efficient way by training a few parameters in the prompt. 
Within the CLIP model, the textual prompts are manually crafted using class labels $y \in \{1, 2, \cdots, K\}$ (e.g., ``a photo of a $\langle$classname$\rangle$") representing $K$ classes. These textual prompts are then tokenized and projected into word embeddings $W=\{ w_1, w_2, \cdots, w_L\} \in \mathbb{R}^{L \times d_l}$ where $L$ is the number of word embeddings and $d_l$ denotes its dimension. To learn the context prompts, we interpose $s (\leq L)$ learnable vectors $\{p_i \in \mathbb{R}^{d_l} \}_{i=1}^s$ in the language branch. Consequently, the textual prompt can be formulated as $P_{k}=\{w_1, p_1, \cdots, p_s, w_{s+2} \cdots, w_L\} \in \mathbb{R}^{L \times d_l}$, where we use $\{p_1,\cdots,p_s\}$ in place of $\{w_2,\cdots,w_{s+1}\}$ to be consistent with previous works. Denote the fixed text encoder as $h(\cdot)$ and image encoder as $g(\cdot)$, and the prediction probabilities for each category are computed with the input prompt $P_{k}$ of class $k$ and image $x$ through matching scores:
\begin{equation}\label{eq1}
    \centering
    q(y=k|\bm{x})=\frac{\exp(sim(g(x),h(P_{k}))/ \tau)}{\sum_{c=1}^{K}\exp(sim(g(x),h(P_{c}))/ \tau)},
\end{equation}
where $sim(\cdot,\cdot)$ denotes a metric function such as cosine similarity, and $\tau$ represents the temperature of Softmax. Next, we optimize the learnable parameters $\{p_i \}_{i=1}^s$ by minimizing the cross-entropy loss 
\begin{equation}\label{eq2}
    \centering
    \ell_{CE} = - \frac{1}{|\mathcal{X}|} \sum_{\bm{x} \in \mathcal{X}} \sum_{k=1}^K y_{\bm{x},k}q(y=k | \bm{x}),
\end{equation} 
where $y_{\bm{x}}$ is a a one-hot label vector.

\subsection{Optimal Transport}
Optimal Transport is a constrained optimization problem aiming to efficiently transfer probability mass between two distributions. Here we briefly recall its formulation of the discrete situation. Given two probability simplex vectors $\alpha$ and $\beta$ and a cost matrix $C \in \mathbb{R}^{|\alpha| \times |\beta|}$, OT aims to find the optimal transport plan $T$ by minimizing the following objective:
\begin{equation}\label{eq3}
    \begin{aligned}
    \centering
    & d_{C}(\alpha, \beta) = \mathop{\min}\limits_{T \in U(\alpha,\beta)} \left \langle C, T\right\rangle, \\
    U(\alpha,\beta)=\! &\left \{ T \!\in \mathbb{R}_{+}^{|\alpha| \times |\beta|}\! \mid T \mathds{1}_{|\beta|} \!= \alpha , T^\top \mathds{1}_{|\alpha|} \!= \beta \right \}, 
    \end{aligned}
\end{equation}
where $\left \langle \cdot, \cdot \right\rangle$ is Frobenius dot-product, $U(\alpha,\beta)$ denotes the solution space of $T$, and $\mathds{1}_d$ is a $d$-dimensional vector of ones. Directly optimizing the OT problem would be time-consuming. Sinkhorn algorithm \cite{cuturi2013sinkhorn} introduces an entropic regularization term for fast optimization. The regularized  OT formulation can be expressed as: $\mathop{\min}\limits_{T \in U(\alpha,\beta)} \left \langle C, T\right\rangle + \lambda \left \langle T, \log T\right\rangle$, 
where $\lambda \geq 0$ is a hyper-parameter. 
In light of this, the optimal transport plan $T^*$ has been shown to be unique with the form $T^* = \text{diag} (u^{(\Tilde{t})}) \exp(-C / \lambda)\text{diag} (v^{(\Tilde{t})})$, where $\Tilde{t}$ represents the iteration and in each iteration $u^{(\Tilde{t})} = u / (\exp (-C/ \lambda)v^{(\Tilde{t}-1)})$ and $v^{(\Tilde{t})}=v / (\exp (-C/ \lambda)^{\top} u^{(\Tilde{t})})$.
 
\section{Methodology}
\label{sec:methodology}

In this section, we present the design of our FedOTP framework, illustrated in Figure \ref{pipeline}. To achieve a balance between global consensus and local personalization, FedOTP utilizes unbalanced Optimal Transport to strengthen the collaboration between global and local prompts, effectively addressing both label shift and feature shift data heterogeneity.

\subsection{Federated Learning with Global and Local Prompts}
Consider a federated learning scenario involving $N$ clients and a central server, and each client $i$ holds a local dataset $D_i=\{(x_i^{j},y_i^{j})\}_{j=1}^{m_i} (i=1,\cdots,N)$ containing $m_i$ samples. Let $D=\{D_1,D_2,\cdots,D_N\}$ represent the total datasets where each dataset is derived from a distinct data distribution $\mathcal{D}_i$, and let $C_t$ represent the set of selected clients participating at communication round $t$. 

Aiming to reduce communication costs and address data heterogeneity, each client is equipped with a pre-trained CLIP model and a prompt learner in our federated learning setup. Through the prompt learner, every client learns both a shared global prompt and a personalized local prompt, allowing clients to extract more individualized insights while maintaining a degree of consensus among them. 
Specifically, for each client $i$, the prompt $P_i$ comprises a global prompt $P_g$ and a personalized prompt $P_{l,i}$, denoted as $P_i=[P_g, P_{l,i}]$. During each communication round $t$, client $i$ initializes the prompt with $P_i^{t,0}=[P_g^{t-1}, P_{l,i}^{t-1}]$. Then the global and local prompts are jointly updated through gradient descent $P_i^{t,r} = P_i^{t,r-1} - \eta \nabla \mathcal{L}_{\mathcal{D}_i} (P_i^{t,r-1})$ for $R$ iterations locally. After local training, only the updated global prompt $P_{g,i}^{t,R}$ is transmitted to the server for aggregation to learn global consensus among clients, while the personalized prompt is retained locally to capture client-specific category characteristics. 
The process of aggregation can be expressed as:
\begin{equation}\label{eq4}
    \centering
    P_g^t = \sum_{i \in C_t}\frac{m_i}{\sum_{j \in C_t}m_j}P_{g,i}^{t,R}.
\end{equation}
Then the objective function of our FedOTP can be formulated as: 
\begin{equation}\label{eq5}
    \centering
    \mathop{\min}\limits_{P_g,\{P_{l,i}\}_{i=1}^N} \sum_{i=1}^{N} \frac{m_i}{\sum_{j \in C_t}m_j}\mathcal{L}_{\mathcal{D}_i}(P_g,P_{l,i}),
\end{equation}
with $\mathcal{L}_{\mathcal{D}_i}(P_g,P_{l,i}) = \mathbb{E}_{(x_i^{j},y_i^{j})\in {D}_i}\ell(f(P_g,P_{l,i};x_i^{j}),y_i^{j})$, where $f(P_g,P_{l,i};\cdot)$ represents the personalized model for client $i$, and $\ell(\cdot,\cdot)$ denotes the cross-entropy loss function as introduced in Eq. (\ref{eq2}).

\subsection{Global-Local Prompt Cooperation by Unbalanced Optimal Transport}
In this subsection, we introduce the details of the prompt learning process for each client, which leverages unbalanced OT to integrate insights learned from both global and local prompts.
To be specific, as shown in Figure \ref{pipeline}, we initialize prompts $P_g$ and $P_{l,i}$ as $\{w_1, p_1, \cdots, p_s, \cdots, w_L\}$ where $w_i$ represents the word embedding and $p_i$ signifies learnable vectors. With the text encoder $h(\cdot)$, we obtain a global textual feature $H_{k,g} = h(P_{g,k}) \in \mathbb{R}^{d_f}$ and a local textual feature $H_{k,l} = h(P_{l,i,k}) \in \mathbb{R}^{d_f}$ for each class $k$, and the combination of these two features is represented as $H_k=[H_{k,g},H_{k,l}]$ for convenience. Uploading an image $x$ to the image encoder $g(\cdot)$, we derive a set of visual features $G = g(x) \in \mathbb{R}^{(V+1) \times d_f}$, which consists of a class token $G_c \in \mathbb{R}^{d_f}$ and a feature map $G_m \in \mathbb{R}^{V \times d_f}$. 

We consider learning an optimal transport plan $T$ that aligns both global and local textual features $H_k$ with visual feature map $G_m$. By representing features as samples from discrete distributions, the cost matrix can be represented by the cosine distance between $H_k$ and $G_m$ as $C = 1 - G_m^\top H_k \in \mathbb{R}^{V \times \text{2}}$, then the optimization objective of unbalanced optimal transport is formulated as:
\begin{equation}\label{eq6}
    \begin{aligned}
    \centering
    & d_{C,k}(\alpha, \beta) = \mathop{\min}\limits_{T \in U(\alpha,\beta)} \left \langle C, T\right\rangle, \\
    U(\alpha,\beta)= &\left \{ T \in \mathbb{R}_{+}^{V \times \text{2}} \mid T \mathds{1}_\text{2} \leq \alpha , T^\top \mathds{1}_V = \beta \right \}, \\
    \end{aligned}
\end{equation}
where $\alpha \in \mathbb{R}^V$ and $\beta \in \mathbb{R}^2$ are essentially marginal probability vectors which satisfy $\left \| \alpha \right \|_1 \geq \left \| \beta \right \|_1 = \gamma$ ($\gamma \in [0,1]$). 
The difference between Eq. (\ref{eq6}) and formulation in PLOT \cite{chen2022prompt} lies in their use of classical OT with two hard equality constraints as Eq. (\ref{eq3}). This forces prompts to map to each image patch, potentially causing them to capture some class-irrelevant information from the image and thereby influencing the final results. In contrast, our FedOTP relaxes one of the equality constraints, allowing prompts to concentrate solely on the most relevant image patches rather than the entire content of the image. Additionally, by controlling $\gamma$, FedOTP owns the ability to regulate the mapping size of prompts on the feature map.

For fast optimization, we add an entropic regularization term to Eq. (\ref{eq6}), and the objective function is formulated as follows:
\begin{equation}\label{eq7}
    \centering
    d_{C,k}(\alpha, \beta) = \mathop{\min}\limits_{T \in U(\alpha,\beta)} \left \langle C, T\right\rangle + \lambda \left \langle T, \log T\right\rangle.
\end{equation}
In line with \cite{benamou2015iterative}, we can further reformulate Eq. (\ref{eq7}) as a Kullback-Leibler (KL) projection, and the solution space $U(\alpha,\beta)$ is then defined as the intersection of two convex but not affine sets: 
\begin{equation}\label{eq8}
    \begin{aligned}
    \centering
    & d_{C,k}(\alpha, \beta) = \mathop{\min}\limits_{T \in U(\alpha,\beta)} \lambda \text{KL}(T\mid e^{-C / \lambda}), \\
    & \mathcal{C}_1 \triangleq \left \{  T \in \mathbb{R}_{+}^{V \times \text{2}} \mid T \mathds{1}_\text{2} \leq \alpha \right \}, \\
    & \mathcal{C}_2 \triangleq \left \{T \in \mathbb{R}_{+}^{V \times \text{2}} \mid T^\top \mathds{1}_V = \beta \right \}.
    \end{aligned}
\end{equation}

To solve Eq. (\ref{eq8}), we employ a rapid implementation of Dykstra's algorithm \cite{dykstra1983algorithm} as introduced in \cite{chang2023csot}, which efficient scales iterative KL projection between $\mathcal{C}_1$ and $\mathcal{C}_2$ by leveraging matrix-vector multiplications exclusively. Initializing $Q = \exp(-C/ \lambda)$ and $v^{(0)} = \mathds{1}_2$, a fast optimization solution is achieved within a few iterations as:
\begin{equation}\label{eq9}
    \centering
    T^* = \text{diag}(u^{(\Tilde{t})}) Q \text{diag}(v^{(\Tilde{t})}),
\end{equation}
where $\Tilde{t}$ is the iteration, and in each iteration $u^{(\Tilde{t})}=\min ( \mathds{1}_V / Q_{\alpha} v^{(\Tilde{t}-1)}, \mathds{1}_V)$ and $v^{(\Tilde{t})} = \mathds{1}_2 / Q_{\beta}^\top u^{(\Tilde{t})}$ with $Q_{\alpha} = Q / \text{diag}(\alpha) \mathds{1}_{V \times 2}$ and $Q_{\beta}^\top = Q^\top / \text{diag}(\beta) \mathds{1}_{V \times 2}$. The details of this algorithm are shown in Appendix Section \ref{Dykstra’s_Algorithm}.

By Eq. (\ref{eq9}), we obtain the optimal transport plan $T^*$ and the final Wasserstein distance $d_{C,k}$, then the matching scores in Eq. (\ref{eq1}) is replaced by the following prediction probability:
\begin{equation}\label{eq10}
    \centering
    q(y=k|\bm{x})=\frac{\exp((1-d_{C,k})/ \tau)}{\sum_{c=1}^{K}\exp((1-d_{C,c})/ \tau)}.
\end{equation}
After obtaining $q(y=k|\bm{x})$, we fix the transport plan $T^*$ and optimize learnable vectors $\{p_a \}_{a=1}^s$ in both global and local prompts simultaneously for client $i$ through cross entropy as described in Eq. (\ref{eq2}). Then the global prompt $P_{g,i}$ is sent to the server for aggregation utilizing Eq. (\ref{eq4}) with the local prompt retained locally.
During local training via OT, the final prediction probability of FedOTP is a synthesis of information derived from both the global and the local prompts. This avoids a straightforward addition of the outcomes from the two prompts, fostering a comprehensive and collaborative learning process. Due to page limitation, the algorithm box is deferred to the Appendix Section \ref{Training_Process}. 

\begin{table*}[t]
\renewcommand\arraystretch{1}
\centering
\caption{The results of our FedOTP and the benchmark methods on the Pathological Non-IID setting with non-overlapping over 10 clients.}
\label{result_clip_datasets}
\resizebox{0.85\textwidth}{!}{
\begin{tabular}{lccccc}
\hline
Methods   & Food101  & DTD   & Caltech101   & Flowers102  & OxfordPets     \\ \hline
\multicolumn{6}{l}{\textbf{\textit{Local Training}}}           \\
Zero-Shot CLIP \cite{radford2021learning} & 75.27±0.05  & 40.21±0.12  & 85.14±0.24  & 62.17±0.12  & 84.47±0.10   \\
CoOp \cite{zhou2022learning}             & 82.54±2.42  & 82.69±0.63  & 90.41±0.44   & 88.23±0.76  & 94.52±1.30    \\ \hline
\multicolumn{6}{l}{\textbf{\textit{Prompt-based Federated Learning}}}           \\
PromptFL \cite{guo2023promptfl}         & 74.81±0.64  & 50.46±0.54    & 87.90±0.54    & 73.68±1.58  & 88.17±1.18    \\
PromptFL+FT \cite{Gary2021fine-tuning}  & 77.16±1.56  & 53.74±1.36     & 89.70±0.25    & 72.31±0.91  & 91.23±0.50   \\
PromptFL+FedProx \cite{li2020federated} & 73.96±0.75  & 50.89±0.71     & 87.80±1.10    & 74.14±0.65  & 87.25±1.48    \\
PromptFL+FedPer \cite{arivazhagan2019federated} 
                                        & 71.29±1.87  & 50.23±0.82     & 86.72±1.45    & 72.11±1.35  & 89.50±1.62    \\
PromptFL+FedAMP \cite{huang2021personalized} & 74.48±1.71  & 47.16±0.92   & 87.31±1.60 & 69.10±0.13 & 80.21±0.44    \\
pFedPrompt \cite{guo2023pfedprompt}     & 92.26±1.34 & 77.14±0.09   & 96.54±1.31    & 86.46±0.15   & 91.84±0.41    \\ 
FedOTP (Ours)                  &  \textbf{92.73±0.15}  & \textbf{87.67±0.70}   &  \textbf{97.02±0.36} & \textbf{96.23±0.44} &  \textbf{98.82±0.11}  \\ \hline
\end{tabular}
}
\end{table*}

\subsection{Generalization Bound}
We analyze the generalization bound of our FedOTP in this section. Before starting the analysis, we first introduce some assumptions as follows. 

\newtheorem{assumption}{Assumption}

\begin{assumption}[Lipschitz Conditions]
\label{assum1}
Let $\mathcal{D}_1, \cdots, \mathcal{D}_N$ denote the real data distribution of each client and $\mathcal{L}_{\mathcal{D}_i}(P_g,P_{l,i}) = \mathbb{E}_{(x_i^{j}, y_i^{j}) \in {D}_i} \ell (f(P_g,P_{l,i}; x_i^{j}), y_i^{j})$ be the expected loss. We assume the following Lipschitz conditions hold:
\begin{subequations}
\begin{equation}
\begin{aligned}
    &| \ell (f((P;x),y) - \ell (f((P^{\prime};x),y)| \\
    & \leq L \|f((P;x),y)- f((P^{\prime};x),y) \|,
\end{aligned}
\end{equation}
\begin{equation}
    \| f(P_g,P_{l,i}) - f(P_g^{\prime},P_{l,i}) \| \leq L_g \|P_g - P_g^{\prime} \|,
\end{equation}
\begin{equation}
    \| f(P_g,P_{l,i}) - f(P_g,P_{l,i}^{\prime}) \| \leq L_{l,i} \|P_{l,i} - P_{l,i}^{\prime} \|.
\end{equation}
\end{subequations}
\end{assumption}

\begin{assumption}
\label{assum2}
Since the convergence of global prompt has been proved in \cite{guo2023promptfl}, we assume $\| \hat{P}_g - P^\ast_g \|_2 \leq A_g$ for convenience. And we assume local prompts $P_{l,i}$ are bounded in a ball of radius $A_{l,i}$, which can be denoted as $\| \hat{P}_{l,i} - P^\ast_{l,i} \|_2 \le A_{l,i}$.
\end{assumption}

Leveraging above assumptions, we can derive the following theorem:

\newtheorem{theorem}{Theorem}
\begin{theorem}[Generalization Bound of FedOTP]
\label{theorem1}
Suppose $\mathcal{\hat{D}}_1,\cdots,\mathcal{\hat{D}}_N$ denote empirical data distribution of $N$ clients with learned parameters $\hat{P}_g$ and $\hat{P}_{l,i}$, and $P_g^\ast$ and $P_{l,i}^\ast$ are optimal parameters for the real distribution $\mathcal{D}_1,\cdots,\mathcal{D}_N$. Let $\mathcal{H}$ represent the personalized hypothesis and $d$ denote the VC-dimension of $\mathcal{H}$. Suppose all the clients participate at every communication round and Assumptions \ref{assum1} and \ref{assum2} hold, with probability at least $1 - \delta$, we have
\begin{equation}
    \begin{aligned}
    \centering
    & \left|\sum_{i=1}^N \frac{m_i}{M}\left(\mathcal{L}_{\hat{\mathcal{D}}_i}(\hat{P}_g,\hat{P}_{l,i})-\mathcal{L}_{\mathcal{D}_i}(P_g^\ast,P_{l,i}^\ast) \right)\right|\\ 
    & \leq\! \sqrt{\frac{M}{2} \!\log\frac{N}{\delta}}  \!+\! \sqrt{\frac{dN}{M} \!\log\frac{eM}{d}} \!+\!  L( L_g A_g \!+\! L_lA_l),
    \end{aligned}
\end{equation}
\end{theorem}
where $M=\sum_{i=1}^N m_i$, and we denote $L_l=\sqrt{\sum_{i=1}^N  L^2_{l,i}}$ and $A_l=\sqrt{\sum_{i=1}^N  A^2_{l,i}}$ for simplicity. Theorem \ref{theorem1} indicates that the performance of FedOTP trained on the empirical distribution relates to the model complexity and Lipschitz assumptions. 
More details and proof of Theorem \ref{theorem1} are provided in the Appendix Section \ref{sec:Generalization_Bound}.

\section{Experiments}
\label{sec:experiments}
In this section, we conducted comprehensive experiments to numerically evaluate our FedOTP in the scenarios of heterogeneous data distribution.

\subsection{Experimental Setup}

\noindent \textbf{Datasets and Data Heterogeneity.} 
We evaluated the performance of FedOTP on nine public benchmark datasets with different types of heterogeneity, including label shift and feature shift. To investigate label shift, we selected two types of datasets. Following previous research \cite{guo2023promptfl, guo2023pfedprompt}, we utilized five visual classification datasets to simulate datasets with limited samples: Food101 \cite{bossard2014food}, DTD \cite{cimpoi2014describing}, Caltech101 \cite{fei2004learning}, Flowers102 \cite{nilsback2008automated}, and OxfordPets \cite{parkhi2012cats}. Referring to these datasets as the CLIP dataset for convenience, we utilized a Pathological setting by randomly allocating a distinct number of non-overlapping classes to each client. 
We also employed two image benchmark datasets: CIFAR-10, and CIFAR-100 \cite{krizhevsky2009learning}. We considered the Dirichlet distribution as introduced in \cite{shamsian2021personalized, li2023fedtp, cao2023knowledge} where datasets are partitioned randomly among clients using a symmetric Dirichlet distribution with $\alpha=0.3$. 
For feature shift, we utilized two datasets with multiple domains: DomainNet \cite{peng2019moment} with 6 domains, and Office-Caltech10 \cite{gong2012geodesic} with 4 domains. In line with prior studies \cite{li2020fedbn,tan2022federated}, each client participating in the federated learning system is assigned data from one of these distinct domains. 
The details of these dataset setup can be found in the Appendix Section \ref{Dataset_Setup}. 

\noindent \textbf{Baselines.} 
We compared FedOTP with three kinds of baselines: (1) Local training methods: (i) Zero-shot CLIP \cite{radford2021learning} with hand-crafted text prompt templates; (ii) CoOp \cite{zhou2022learning} with learnable prompt vectors trained on each client locally. (2) Existing prompt-based federated learning methods: (i) PromptFL \cite{guo2023promptfl} learning a unified prompt across clients; (ii) pFedPrompt \cite{guo2023pfedprompt} learning a shared prompt with personalized visual attention modules. (3) Four adapted methods derived from traditional PFL techniques, including PromptFL + FT \cite{Gary2021fine-tuning}, PromptFL+FedProx \cite{li2020federated}, PromptFL+FedPer \cite{arivazhagan2019federated} and PromptFL+FedAMP \cite{huang2021personalized}, as introduced in \cite{guo2023pfedprompt}.

\noindent \textbf{Implementation Details.} 
To simulate federated learning in various scenarios, we consider the following two settings: (1) $n=10$ clients with a full $100\%$ partition, (2) $n=100$ clients with a $10\%$ partition. 
We employ SGD optimizer with a learning rate $lr=0.001$ and local epoch $R=5$ for CLIP datasets while $R=1$ for other cases. The communication round is set to $T=10$ for CLIP datasets with $10$ clients and $T=150$ for CIFAR-10/CIFAR-100 datasets with $100$ clients. We present the results using two representative backbones, ResNet50 \cite{he2016deep} and ViT\_B16 \cite{dosovitskiy2020image}, defaulting to ViT\_B16 if not explicitly specified. More implementation details can be found in the Appendix Section \ref{Implementation_Details}.

\begin{table*}[t]
\renewcommand\arraystretch{1}
\centering
\caption{Experimental results on DomainNet dataset with feature $\&$ label shifts.}
\label{result_domain_datasets_2}
\resizebox{0.98\textwidth}{!}{
\begin{tabular}{lccccccc}
\hline
Datasets       & \multicolumn{7}{c}{DomainNet}                                                                   \\ \cline{2-8} 
Domains         & Clipart           & Infograph           & Painting           & Quickdraw           & Real           & Sketch           & Avg.        \\ \hline
\multicolumn{6}{l}{\textbf{\textit{Local Training}}}           \\
Zero-Shot CLIP \cite{radford2021learning}          & 8.72±1.73  & 12.48±3.78  & 8.53±4.32   & 9.31±0.69   & 9.13±2.55   & 11.96±2.80  & 10.02±2.65   \\
CoOp \cite{zhou2022learning} & 44.40±14.89 & 45.68±16.53  & \textbf{47.21±18.20} & \textbf{41.13±20.62} & 48.02±24.49 & 39.47±5.68 & 44.32±16.74 \\ \hline
\multicolumn{6}{l}{\textbf{\textit{Prompt-based Federated Learning}}}           \\
PromptFL \cite{guo2023promptfl}      & 9.31±6.53   & 12.58±9.91  & 8.23± 8.47  & 14.79±12.07  & 9.37±10.82  & 7.48±11.32  & 10.29±10.35 \\
PromptFL+FedProx \cite{li2020federated}    & 9.84±6.60 & 11.16±11.17 & 10.64±6.79   & 13.40±16.09  & 9.39±7.69   & 6.78±11.76  & 10.20±10.99 \\ 
FedOTP (Ours)           & \textbf{46.14±6.53} & \textbf{60.14±18.23}  & 45.2±16.86  & 38.66±7.60  & \textbf{49.30±17.80} & \textbf{49.02±24.22}  & \textbf{48.08±15.21} \\ \hline
\end{tabular}
}
\end{table*}

\begin{figure*}[ht]
\centering
\includegraphics[width=1\textwidth]{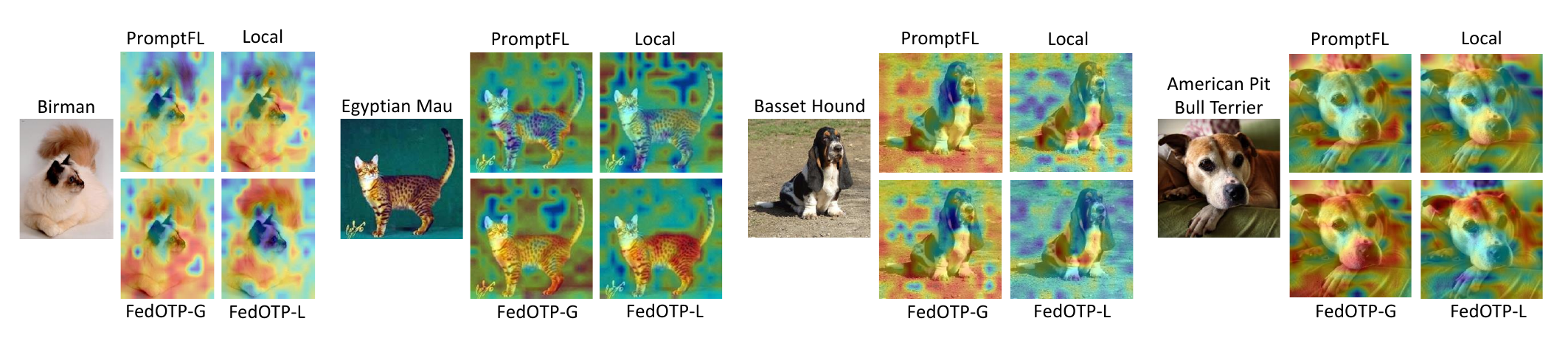} 
\caption{Heatmaps of similarity between text features and image feature maps for different methods on 4 categories in OxfordPets dataset. ``FedOTP-G'' denotes the results from the global prompt and ``FedOTP-L'' refers to the local prompt.}
\label{heatmap}
\end{figure*}

\subsection{Performance Evaluation}
\noindent \textbf{Evaluation Protocol.} 
We evaluated the models on each client's private test data whose distribution is consistent with its training set. The reported results are the average test accuracy across all clients from three different seeds.

\begin{table}[h]
\renewcommand\arraystretch{1}
\centering
\caption{The results of our FedOTP and the benchmark methods on Dirichlet settings in CIFAR-10 and CIFAR-100 over 100 clients.}
\label{result_cifar_datasets}
\resizebox{0.88\columnwidth}{!}{
\begin{tabular}{lcc}
\hline  
Methods      & CIFAR-10              & CIFAR-100              \\ \hline
\multicolumn{3}{l}{\textbf{\textit{Local Training}}}           \\
Zero-Shot CLIP \cite{radford2021learning}          & 87.71±0.68                & 64.92±0.53          \\
CoOp \cite{zhou2022learning}      & 93.11±0.39                                 &  74.83±0.45                   \\ \hline
\multicolumn{3}{l}{\textbf{\textit{Prompt-based Federated Learning}}}           \\
PromptFL \cite{guo2023promptfl}    & 92.30±0.87          & 73.67±0.56          \\
PromptFL+FedProx \cite{li2020federated}    & 91.83±0.47           & 71.11±0.91          \\
FedOTP (Ours)         & \textbf{96.05±0.12}   &  \textbf{78.03±0.08} \\ \hline
\end{tabular}}
\end{table}

\begin{figure*}[th]
\centering
\includegraphics[width=1\textwidth]{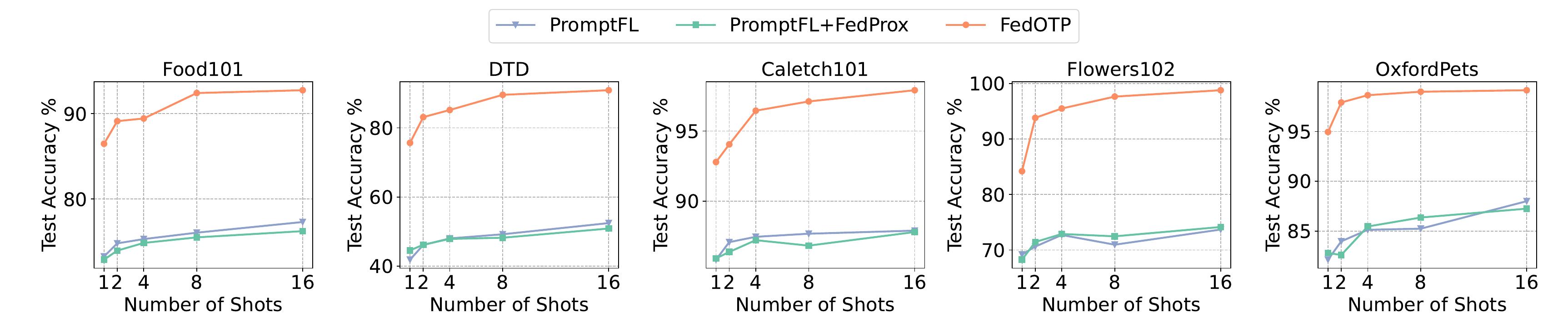} 
\caption{Performance with the different number of shots. }
\label{few_shot}
\end{figure*}

\begin{table*}[th]
\renewcommand\arraystretch{1.1}
\tabcolsep=0.15cm
\centering
\caption{Quantitative comparisons on the Pathological Non-IID setting across different numbers of shots over 10 clients.}
\label{result_2&8shot}
\resizebox{1\textwidth}{!}{
\begin{tabular}{lcccccccccc}
\hline
Datasets                      & \multicolumn{2}{c}{Food101} & \multicolumn{2}{c}{DTD} & \multicolumn{2}{c}{Caltech101} & \multicolumn{2}{c}{Flowers102} & \multicolumn{2}{c}{OxfordPets} \\ \cline{2-11} 
Number of shots                   & 2            & 8            & 2          & 8          & 2              & 8             & 2              & 8             & 2              & 8             \\ \hline
FedOTP (Similarity Averaging) & 83.38±0.54   & 87.59±1.05   & 81.01±0.23 & 88.17±0.73 & 92.68±0.44     & 96.73±0.29    & 91.73±0.68     & 97.09±0.18    & 96.23±0.25     & 98.34±0.15    \\
FedOTP (Classical OT)         & 88.07±0.63   & 89.77±0.62   & 81.42±0.99 & 88.43±0.45 & 93.17±0.68     & 96.80±0.23    & 92.84±1.34     & 97.07±0.25    & 96.55±0.26     & 98.51±0.27    \\
FedOTP (Unbalanced OT)        & \textbf{89.12±0.28}   & \textbf{92.94±0.18}   & \textbf{85.50±0.35} & \textbf{90.25±0.74} & \textbf{95.05±0.49}     & \textbf{97.34±0.18}    & \textbf{93.96±0.48}     & \textbf{98.23±0.32}    & \textbf{97.73±0.57}     & \textbf{99.02±0.38}    \\ \hline
\end{tabular}
}
\end{table*}

\noindent \textbf{Model Evaluation on Label Shifts.} 
We first measured the performance of FedOTP against baselines on datasets with label shifts. The experimental results on CLIP datasets and CIFAR-10/CIFAR-100 datasets are summarized in Table \ref{result_clip_datasets} and Table \ref{result_cifar_datasets}.  For easy comparison, Table \ref{result_clip_datasets} reports results utilizing ResNet50 as the backbone, maintaining consistency with \cite{guo2023pfedprompt}. As shown in Table \ref{result_clip_datasets}, our FedOTP outperforms state-of-the-art algorithms by a large margin across all datasets, which confirms the effectiveness of our Global-Local prompt cooperation mechanism to handle label shift scenarios. 
Remarkably, while both PromptFL+FedPer (which splits the learnable prompt vector into \textit{``base+personalized"} vectors) and pFedPrompt (utilizing a shared prompt with a personalized attention module in the vision modal) experience significant declines when datasets are altered, FedOTP exhibits slight fluctuations. This verifies the robustness of our method across diverse scenarios.
Table \ref{result_cifar_datasets} shows the results of our FedOTP and benchmark methods on CIFAR-10/CIFAR-100 datasets under Dirichlet setting over $100$ clients with $10\%$ partition. Even in this scenario with Dirichlet settings and a large number of clients, FedOTP consistently outperforms the baseline methods, further highlighting the superiority of our approach.

\noindent \textbf{Model Evaluation on Feature $\&$ Label Shifts.} 
In this set of experiments, we explored scenarios involving both feature shifts and label shifts by partitioning data within a domain into five clients based on the Dirichlet distribution with $\alpha=0.1$. We analyzed the mean and variance of clients in the same domain, and the outcomes for the DomainNet dataset are summarized in Table \ref{result_domain_datasets_2}.
In the presence of two types of data heterogeneity, our method performs favorably against baselines. We observe that, with significant data heterogeneity across clients, traditional federated learning methods experience a pronounced performance decline compared to local training. In contrast, our FedOTP exhibits superior performance, achieving a $3.7\%$ increase in average accuracy on each domain. Additional experimental results on feature shifts and in the Office-Caltech10 dataset are available in the Appendix Section \ref{Model_Evaluation_on_Feature_Shifts} and \ref{Feature_and_Label_Shifts}.

\noindent \textbf{Visualization.} 
We first investigated the interplay between global and local prompts by representing the similarity between text features and image feature maps as a heatmap on OxfordPets \cite{parkhi2012cats} dataset.
To be specific, we compared the heatmaps of our FedOTP with PromptFL and Local Training using CoOp, and the original images and corresponding heatmaps are illustrated in Figure \ref{heatmap}. We observed that global prompts of FedOTP might concentrate more on common features, like limbs and facial characteristics, while local prompts tended to capture client-specific details such as the special tail of ``Birman", unique patterns of ``Egyptian Mau" and ``Basset Hound", and the distinct dent on the head of ``Terrier". 
This demonstrates the effectiveness of FedOTP in balancing global consensus and local personalization. 
More visualization results on transport plans of our FedOTP will be given in the Appendix Section \ref{Visualizations_of_Transport_Plans}.

\subsection{Ablation Study}

\noindent \textbf{Impact of Number of Shots.} 
Following the few-shot evaluation setting adopted in \cite{guo2023promptfl,guo2023pfedprompt}, we further investigated the impact of the number of shots in FedOTP. To analyze this, we varied the number of shots during the training process within the range of [1, 2, 4, 8, 16]. Results are summarized in Figure \ref{few_shot}, where the horizontal axis denotes the number of shots and the vertical axis represents the average test accuracy. We observe that as the number of shots increases, the corresponding performance of each method gradually improves. However, our FedOTP consistently exhibits a dominant edge over methods with a shared global prompt in all scenarios. 

\noindent \textbf{Effectiveness of the Unbalanced OT.}
In this subsection, we explored the effectiveness of OT on two variants of FedOTP briefly described below: (1) FedOTP (Similarity Averaging): removing OT in FedOTP and matching global and local prompts with visual feature maps by averaging similarities of each visual-textual pair; (2) FedOTP (Classical OT): employing classical OT during the matching process. 
The results in Table \ref{result_2&8shot} demonstrate the effectiveness of utilizing OT to align feature maps with global and local prompts compared to FedOTP (Similarity Averaging) in almost all cases, particularly on the Food101 dataset.
This is because the absence of OT leads to the feature map's distance from prompts reverting to the mean distance of each feature-prompt pair, highlighting the crucial role of OT in providing resilience to visual misalignment. 
In addition, the persistent superiority of unbalanced OT over classical OT across all scenarios serves as a compelling testament to the effectiveness of our approach.

\section{Conclusion}
\label{sec:conclusion}
In this paper, we proposed Federated Prompts Cooperation via Optimal Transport (FedOTP), a novel framework designed to facilitate efficient model personalization across heterogeneous clients. In our approach, each client is equipped with both a global prompt and a local prompt, and then unbalanced Optimal Transport is utilized to align local visual features with these prompts, fostering enhanced collaboration between global and local prompts. 
With fine-grained matching facilitated by OT, FedOTP effectively addresses data heterogeneity characterized by domain discrepancy and imbalanced class distributions. 
Our extensive experiments across diverse datasets consistently demonstrate the superior performance of FedOTP in tackling both label shifts and feature shifts, which verifies the effectiveness of our Global-Local prompt cooperation mechanism via OT. Through visualization results, we confirmed that global prompts learned by FedOTP concentrated on common features among all clients, while local prompts captured individual client-specific details. In future work, we aim to investigate the generalization capabilities of our method on novel clients unseen during the training process.

\section*{Acknowledgement}
This work was supported by NSFC (No.62303319), Shanghai Sailing Program (22YF1428800, 21YF1429400), Shanghai Local College Capacity Building Program (23010503100), Shanghai Frontiers Science Center of Human-centered Artificial Intelligence (ShangHAI), MoE Key Laboratory of Intelligent Perception and Human-Machine Collaboration (ShanghaiTech University), and Shanghai Engineering Research Center of Intelligent Vision and Imaging.

{
    \small
    \bibliographystyle{ieeenat_fullname}
    \bibliography{reference}

\begin{thebibliography}{78}
\providecommand{\natexlab}[1]{#1}
\providecommand{\url}[1]{\texttt{#1}}
\expandafter\ifx\csname urlstyle\endcsname\relax
  \providecommand{\doi}[1]{doi: #1}\else
  \providecommand{\doi}{doi: \begingroup \urlstyle{rm}\Url}\fi

\bibitem[Arivazhagan et~al.(2019)Arivazhagan, Aggarwal, Singh, and
  Choudhary]{arivazhagan2019federated}
Manoj~Ghuhan Arivazhagan, Vinay Aggarwal, Aaditya~Kumar Singh, and Sunav
  Choudhary.
\newblock Federated learning with personalization layers.
\newblock \emph{arXiv preprint arXiv:1912.00818}, 2019.

\bibitem[Benamou et~al.(2015)Benamou, Carlier, Cuturi, Nenna, and
  Peyr{\'e}]{benamou2015iterative}
Jean-David Benamou, Guillaume Carlier, Marco Cuturi, Luca Nenna, and Gabriel
  Peyr{\'e}.
\newblock Iterative bregman projections for regularized transportation
  problems.
\newblock \emph{SIAM Journal on Scientific Computing}, 37\penalty0
  (2):\penalty0 A1111--A1138, 2015.

\bibitem[Bossard et~al.(2014)Bossard, Guillaumin, and
  Van~Gool]{bossard2014food}
Lukas Bossard, Matthieu Guillaumin, and Luc Van~Gool.
\newblock Food-101--mining discriminative components with random forests.
\newblock In \emph{Computer Vision--ECCV 2014: 13th European Conference,
  Zurich, Switzerland, September 6-12, 2014, Proceedings, Part VI 13}, pages
  446--461. Springer, 2014.

\bibitem[Cao et~al.(2023)Cao, Shi, Yu, Wang, and Tao]{cao2023knowledge}
Yu-Tong Cao, Ye Shi, Baosheng Yu, Jingya Wang, and Dacheng Tao.
\newblock Knowledge-aware federated active learning with non-iid data.
\newblock In \emph{Proceedings of the IEEE/CVF International Conference on
  Computer Vision}, pages 22279--22289, 2023.

\bibitem[Chang et~al.(2022)Chang, Shi, Tuan, and Wang]{chang2022unified}
Wanxing Chang, Ye Shi, Hoang Tuan, and Jingya Wang.
\newblock Unified optimal transport framework for universal domain adaptation.
\newblock \emph{Advances in Neural Information Processing Systems},
  35:\penalty0 29512--29524, 2022.

\bibitem[Chang et~al.(2023)Chang, Shi, and Wang]{chang2023csot}
Wanxing Chang, Ye Shi, and Jingya Wang.
\newblock {CSOT: Curriculum and Structure-Aware Optimal Transport for Learning
  with Noisy Labels}.
\newblock In \emph{Thirty-seventh Conference on Neural Information Processing
  Systems}, 2023.

\bibitem[Chen et~al.(2022)Chen, Yao, Song, Li, Rao, and Zhang]{chen2022prompt}
Guangyi Chen, Weiran Yao, Xiangchen Song, Xinyue Li, Yongming Rao, and Kun
  Zhang.
\newblock Prompt learning with optimal transport for vision-language models.
\newblock \emph{arXiv preprint arXiv:2210.01253}, 2022.

\bibitem[Chen and Chao(2021)]{chen2021bridging}
Hong-You Chen and Wei-Lun Chao.
\newblock On bridging generic and personalized federated learning for image
  classification.
\newblock \emph{arXiv preprint arXiv:2107.00778}, 2021.

\bibitem[Chiang et~al.(2023)Chiang, Terai, Chiang, Lin, Ji, and
  Lui]{chiang2023optimal}
Yi-Han Chiang, Koudai Terai, Tsung-Wei Chiang, Hai Lin, Yusheng Ji, and John~CS
  Lui.
\newblock {Optimal Transport based One-Shot Federated Learning for Artificial
  Intelligence of Things}.
\newblock \emph{IEEE Internet of Things Journal}, 2023.

\bibitem[Chizat et~al.(2018)Chizat, Peyr{\'e}, Schmitzer, and
  Vialard]{chizat2018scaling}
Lenaic Chizat, Gabriel Peyr{\'e}, Bernhard Schmitzer, and Fran{\c{c}}ois-Xavier
  Vialard.
\newblock Scaling algorithms for unbalanced optimal transport problems.
\newblock \emph{Mathematics of Computation}, 87\penalty0 (314):\penalty0
  2563--2609, 2018.

\bibitem[Cimpoi et~al.(2014)Cimpoi, Maji, Kokkinos, Mohamed, and
  Vedaldi]{cimpoi2014describing}
Mircea Cimpoi, Subhransu Maji, Iasonas Kokkinos, Sammy Mohamed, and Andrea
  Vedaldi.
\newblock Describing textures in the wild.
\newblock In \emph{Proceedings of the IEEE conference on computer vision and
  pattern recognition}, pages 3606--3613, 2014.

\bibitem[Collins et~al.(2021)Collins, Hassani, Mokhtari, and
  Shakkottai]{collins2021exploiting}
Liam Collins, Hamed Hassani, Aryan Mokhtari, and Sanjay Shakkottai.
\newblock Exploiting shared representations for personalized federated
  learning.
\newblock In \emph{International Conference on Machine Learning}, pages
  2089--2099. PMLR, 2021.

\bibitem[Cuturi(2013)]{cuturi2013sinkhorn}
Marco Cuturi.
\newblock {Sinkhorn distances: Lightspeed computation of optimal transport}.
\newblock \emph{Advances in neural information processing systems}, 26, 2013.

\bibitem[Damodaran et~al.(2018)Damodaran, Kellenberger, Flamary, Tuia, and
  Courty]{damodaran2018deepjdot}
Bharath~Bhushan Damodaran, Benjamin Kellenberger, R{\'e}mi Flamary, Devis Tuia,
  and Nicolas Courty.
\newblock {Deepjdot: Deep joint distribution optimal transport for unsupervised
  domain adaptation}.
\newblock In \emph{Proceedings of the European conference on computer vision
  (ECCV)}, pages 447--463, 2018.

\bibitem[Dosovitskiy et~al.(2020)Dosovitskiy, Beyer, Kolesnikov, Weissenborn,
  Zhai, Unterthiner, Dehghani, Minderer, Heigold, Gelly,
  et~al.]{dosovitskiy2020image}
Alexey Dosovitskiy, Lucas Beyer, Alexander Kolesnikov, Dirk Weissenborn,
  Xiaohua Zhai, Thomas Unterthiner, Mostafa Dehghani, Matthias Minderer, Georg
  Heigold, Sylvain Gelly, et~al.
\newblock {An image is worth 16x16 words: Transformers for image recognition at
  scale}.
\newblock \emph{arXiv preprint arXiv:2010.11929}, 2020.

\bibitem[Dykstra(1983)]{dykstra1983algorithm}
Richard~L Dykstra.
\newblock An algorithm for restricted least squares regression.
\newblock \emph{Journal of the American Statistical Association}, 78\penalty0
  (384):\penalty0 837--842, 1983.

\bibitem[Fallah et~al.(2020)Fallah, Mokhtari, and
  Ozdaglar]{fallah2020personalized}
Alireza Fallah, Aryan Mokhtari, and Asuman Ozdaglar.
\newblock {Personalized federated learning with theoretical guarantees: A
  model-agnostic meta-learning approach}.
\newblock \emph{Advances in Neural Information Processing Systems},
  33:\penalty0 3557--3568, 2020.

\bibitem[Farnia et~al.(2022)Farnia, Reisizadeh, Pedarsani, and
  Jadbabaie]{farnia2022optimal}
Farzan Farnia, Amirhossein Reisizadeh, Ramtin Pedarsani, and Ali Jadbabaie.
\newblock An optimal transport approach to personalized federated learning.
\newblock \emph{IEEE Journal on Selected Areas in Information Theory},
  3\penalty0 (2):\penalty0 162--171, 2022.

\bibitem[Fatras et~al.(2021)Fatras, S{\'e}journ{\'e}, Flamary, and
  Courty]{fatras2021unbalanced}
Kilian Fatras, Thibault S{\'e}journ{\'e}, R{\'e}mi Flamary, and Nicolas Courty.
\newblock Unbalanced minibatch optimal transport; applications to domain
  adaptation.
\newblock In \emph{International Conference on Machine Learning}, pages
  3186--3197. PMLR, 2021.

\bibitem[Fei-Fei et~al.(2004)Fei-Fei, Fergus, and Perona]{fei2004learning}
Li Fei-Fei, Rob Fergus, and Pietro Perona.
\newblock {Learning generative visual models from few training examples: An
  incremental bayesian approach tested on 101 object categories}.
\newblock In \emph{2004 conference on computer vision and pattern recognition
  workshop}, pages 178--178. IEEE, 2004.

\bibitem[Feng et~al.(2023{\natexlab{a}})Feng, Ren, and Xie]{feng2023ot}
Chuanwen Feng, Yilong Ren, and Xike Xie.
\newblock {OT-Filter: An Optimal Transport Filter for Learning With Noisy
  Labels}.
\newblock In \emph{Proceedings of the IEEE/CVF Conference on Computer Vision
  and Pattern Recognition}, pages 16164--16174, 2023{\natexlab{a}}.

\bibitem[Feng et~al.(2023{\natexlab{b}})Feng, Li, Xu, Liu, Fu, and
  Zuo]{feng2023learning}
Chun-Mei Feng, Bangjun Li, Xinxing Xu, Yong Liu, Huazhu Fu, and Wangmeng Zuo.
\newblock {Learning Federated Visual Prompt in Null Space for MRI
  Reconstruction}.
\newblock In \emph{Proceedings of the IEEE/CVF Conference on Computer Vision
  and Pattern Recognition}, pages 8064--8073, 2023{\natexlab{b}}.

\bibitem[Frogner et~al.(2015)Frogner, Zhang, Mobahi, Araya, and
  Poggio]{frogner2015learning}
Charlie Frogner, Chiyuan Zhang, Hossein Mobahi, Mauricio Araya, and Tomaso~A
  Poggio.
\newblock {Learning with a Wasserstein loss}.
\newblock \emph{Advances in neural information processing systems}, 28, 2015.

\bibitem[Gary~Cheng and Duchi(2021)]{Gary2021fine-tuning}
Karan N.~Chadha Gary~Cheng and John~C. Duchi.
\newblock {Fine-tuning is Fine in Federated Learning}.
\newblock \emph{CoRR}, abs/2108.07313, 2021.

\bibitem[Gong et~al.(2012)Gong, Shi, Sha, and Grauman]{gong2012geodesic}
Boqing Gong, Yuan Shi, Fei Sha, and Kristen Grauman.
\newblock Geodesic flow kernel for unsupervised domain adaptation.
\newblock In \emph{2012 IEEE conference on computer vision and pattern
  recognition}, pages 2066--2073. IEEE, 2012.

\bibitem[Guo et~al.(2023{\natexlab{a}})Guo, Guo, and Wang]{guo2023pfedprompt}
Tao Guo, Song Guo, and Junxiao Wang.
\newblock {pFedPrompt: Learning Personalized Prompt for Vision-Language Models
  in Federated Learning}.
\newblock In \emph{Proceedings of the ACM Web Conference 2023}, pages
  1364--1374, 2023{\natexlab{a}}.

\bibitem[Guo et~al.(2023{\natexlab{b}})Guo, Guo, Wang, Tang, and
  Xu]{guo2023promptfl}
Tao Guo, Song Guo, Junxiao Wang, Xueyang Tang, and Wenchao Xu.
\newblock {PromptFL: Let federated participants cooperatively learn prompts
  instead of models-federated learning in age of foundation model}.
\newblock \emph{IEEE Transactions on Mobile Computing}, 2023{\natexlab{b}}.

\bibitem[Hanzely and Richt{\'a}rik(2020)]{hanzely2020federated}
Filip Hanzely and Peter Richt{\'a}rik.
\newblock Federated learning of a mixture of global and local models.
\newblock \emph{arXiv preprint arXiv:2002.05516}, 2020.

\bibitem[He et~al.(2016)He, Zhang, Ren, and Sun]{he2016deep}
Kaiming He, Xiangyu Zhang, Shaoqing Ren, and Jian Sun.
\newblock Deep residual learning for image recognition.
\newblock In \emph{Proceedings of the IEEE conference on computer vision and
  pattern recognition}, pages 770--778, 2016.

\bibitem[He et~al.(2022)He, Zheng, Tay, Gupta, Du, Aribandi, Zhao, Li, Chen,
  Metzler, et~al.]{he2022hyperprompt}
Yun He, Steven Zheng, Yi Tay, Jai Gupta, Yu Du, Vamsi Aribandi, Zhe Zhao,
  YaGuang Li, Zhao Chen, Donald Metzler, et~al.
\newblock {Hyperprompt: Prompt-based task-conditioning of transformers}.
\newblock In \emph{International Conference on Machine Learning}, pages
  8678--8690. PMLR, 2022.

\bibitem[Huang et~al.(2023)Huang, Shi, Cai, and Suzuki]{huang2023understanding}
Wei Huang, Ye Shi, Zhongyi Cai, and Taiji Suzuki.
\newblock Understanding convergence and generalization in federated learning
  through feature learning theory.
\newblock In \emph{The Twelfth International Conference on Learning
  Representations}, 2023.

\bibitem[Huang et~al.(2021)Huang, Chu, Zhou, Wang, Liu, Pei, and
  Zhang]{huang2021personalized}
Yutao Huang, Lingyang Chu, Zirui Zhou, Lanjun Wang, Jiangchuan Liu, Jian Pei,
  and Yong Zhang.
\newblock {Personalized Cross-Silo Federated Learning on Non-IID Data}.
\newblock In \emph{AAAI}, pages 7865--7873, 2021.

\bibitem[Kantorovich(2006)]{kantorovich2006translocation}
Leonid~V Kantorovich.
\newblock On the translocation of masses.
\newblock \emph{Journal of mathematical sciences}, 133\penalty0 (4):\penalty0
  1381--1382, 2006.

\bibitem[Khattak et~al.(2023)Khattak, Rasheed, Maaz, Khan, and
  Khan]{khattak2023maple}
Muhammad~Uzair Khattak, Hanoona Rasheed, Muhammad Maaz, Salman Khan, and
  Fahad~Shahbaz Khan.
\newblock {Maple: Multi-modal prompt learning}.
\newblock In \emph{Proceedings of the IEEE/CVF Conference on Computer Vision
  and Pattern Recognition}, pages 19113--19122, 2023.

\bibitem[Khodak et~al.(2019)Khodak, Balcan, and Talwalkar]{khodak2019adaptive}
Mikhail Khodak, Maria-Florina~F Balcan, and Ameet~S Talwalkar.
\newblock Adaptive gradient-based meta-learning methods.
\newblock \emph{Advances in Neural Information Processing Systems}, 32, 2019.

\bibitem[Krizhevsky et~al.(2009)Krizhevsky, Hinton,
  et~al.]{krizhevsky2009learning}
Alex Krizhevsky, Geoffrey Hinton, et~al.
\newblock Learning multiple layers of features from tiny images.
\newblock 2009.

\bibitem[Lahn et~al.(2024)Lahn, Raghvendra, and Zhang]{lahn2024combinatorial}
Nathaniel Lahn, Sharath Raghvendra, and Kaiyi Zhang.
\newblock {A Combinatorial Algorithm for Approximating the Optimal Transport in
  the Parallel and MPC Settings}.
\newblock \emph{Advances in Neural Information Processing Systems}, 36, 2024.

\bibitem[Li et~al.(2024)Li, Shi, Yu, and Wang]{li2024unsupervised}
Bin Li, Ye Shi, Qian Yu, and Jingya Wang.
\newblock Unsupervised cross-domain image retrieval via prototypical optimal
  transport.
\newblock \emph{arXiv preprint arXiv:2402.18411}, 2024.

\bibitem[Li et~al.(2023)Li, Cai, Wang, Tang, Ding, Lin, and Shi]{li2023fedtp}
Hongxia Li, Zhongyi Cai, Jingya Wang, Jiangnan Tang, Weiping Ding, Chin-Teng
  Lin, and Ye Shi.
\newblock {FedTP: Federated Learning by Transformer Personalization}.
\newblock \emph{IEEE Transactions on Neural Networks and Learning Systems},
  2023.

\bibitem[Li et~al.(2021{\natexlab{a}})Li, He, and Song]{li2021model}
Qinbin Li, Bingsheng He, and Dawn Song.
\newblock Model-contrastive federated learning.
\newblock In \emph{Proceedings of the IEEE/CVF conference on computer vision
  and pattern recognition}, pages 10713--10722, 2021{\natexlab{a}}.

\bibitem[Li et~al.(2021{\natexlab{b}})Li, Wang, Liu, Li, and Xu]{li2021causal}
Qian Li, Zhichao Wang, Shaowu Liu, Gang Li, and Guandong Xu.
\newblock Causal optimal transport for treatment effect estimation.
\newblock \emph{IEEE transactions on neural networks and learning systems},
  2021{\natexlab{b}}.

\bibitem[Li et~al.(2020{\natexlab{a}})Li, Sahu, Zaheer, Sanjabi, Talwalkar, and
  Smith]{li2020federated}
Tian Li, Anit~Kumar Sahu, Manzil Zaheer, Maziar Sanjabi, Ameet Talwalkar, and
  Virginia Smith.
\newblock Federated optimization in heterogeneous networks.
\newblock \emph{Proceedings of Machine Learning and Systems}, 2:\penalty0
  429--450, 2020{\natexlab{a}}.

\bibitem[Li et~al.(2021{\natexlab{c}})Li, Hu, Beirami, and Smith]{li2021ditto}
Tian Li, Shengyuan Hu, Ahmad Beirami, and Virginia Smith.
\newblock {Ditto: Fair and robust federated learning through personalization}.
\newblock In \emph{International Conference on Machine Learning}, pages
  6357--6368. PMLR, 2021{\natexlab{c}}.

\bibitem[Li et~al.(2020{\natexlab{b}})Li, JIANG, Zhang, Kamp, and
  Dou]{li2020fedbn}
Xiaoxiao Li, Meirui JIANG, Xiaofei Zhang, Michael Kamp, and Qi Dou.
\newblock {FedBN: Federated Learning on Non-IID Features via Local Batch
  Normalization}.
\newblock In \emph{International Conference on Learning Representations},
  2020{\natexlab{b}}.

\bibitem[Liang et~al.(2020)Liang, Liu, Ziyin, Allen, Auerbach, Brent,
  Salakhutdinov, and Morency]{liang2020think}
Paul~Pu Liang, Terrance Liu, Liu Ziyin, Nicholas~B Allen, Randy~P Auerbach,
  David Brent, Ruslan Salakhutdinov, and Louis-Philippe Morency.
\newblock {Think locally, act globally: Federated learning with local and
  global representations}.
\newblock \emph{arXiv preprint arXiv:2001.01523}, 2020.

\bibitem[Liu et~al.(2023{\natexlab{a}})Liu, Yuan, Fu, Jiang, Hayashi, and
  Neubig]{liu2023pre}
Pengfei Liu, Weizhe Yuan, Jinlan Fu, Zhengbao Jiang, Hiroaki Hayashi, and
  Graham Neubig.
\newblock Pre-train, prompt, and predict: A systematic survey of prompting
  methods in natural language processing.
\newblock \emph{ACM Computing Surveys}, 55\penalty0 (9):\penalty0 1--35,
  2023{\natexlab{a}}.

\bibitem[Liu et~al.(2023{\natexlab{b}})Liu, Lu, Liu, An, Xu, Yao, Zhang, Xiong,
  and Gui]{liu2023hierarchical}
Yajing Liu, Yuning Lu, Hao Liu, Yaozu An, Zhuoran Xu, Zhuokun Yao, Baofeng
  Zhang, Zhiwei Xiong, and Chenguang Gui.
\newblock {Hierarchical Prompt Learning for Multi-Task Learning}.
\newblock In \emph{Proceedings of the IEEE/CVF Conference on Computer Vision
  and Pattern Recognition}, pages 10888--10898, 2023{\natexlab{b}}.

\bibitem[Lu et~al.(2023)Lu, Hu, Wang, and Xie]{lu2023fedclip}
Wang Lu, Xixu Hu, Jindong Wang, and Xing Xie.
\newblock {FedCLIP: Fast Generalization and Personalization for CLIP in
  Federated Learning}.
\newblock \emph{arXiv preprint arXiv:2302.13485}, 2023.

\bibitem[Lu et~al.(2022)Lu, Liu, Zhang, Liu, and Tian]{lu2022prompt}
Yuning Lu, Jianzhuang Liu, Yonggang Zhang, Yajing Liu, and Xinmei Tian.
\newblock Prompt distribution learning.
\newblock In \emph{Proceedings of the IEEE/CVF Conference on Computer Vision
  and Pattern Recognition}, pages 5206--5215, 2022.

\bibitem[Mansour et~al.(2020)Mansour, Mohri, Ro, and Suresh]{mansour2020three}
Yishay Mansour, Mehryar Mohri, Jae Ro, and Ananda~Theertha Suresh.
\newblock Three approaches for personalization with applications to federated
  learning.
\newblock \emph{arXiv preprint arXiv:2002.10619}, 2020.

\bibitem[Marfoq et~al.(2022)Marfoq, Neglia, Vidal, and
  Kameni]{marfoq2022personalized}
Othmane Marfoq, Giovanni Neglia, Richard Vidal, and Laetitia Kameni.
\newblock {Personalized Federated Learning through Local Memorization}.
\newblock In \emph{International Conference on Machine Learning}, pages
  15070--15092. PMLR, 2022.

\bibitem[McMahan et~al.(2017)McMahan, Moore, Ramage, Hampson, and
  y~Arcas]{mcmahan2017communication}
Brendan McMahan, Eider Moore, Daniel Ramage, Seth Hampson, and Blaise~Aguera y
  Arcas.
\newblock Communication-efficient learning of deep networks from decentralized
  data.
\newblock In \emph{Artificial intelligence and statistics}, pages 1273--1282.
  PMLR, 2017.

\bibitem[Mohri et~al.(2018)Mohri, Rostamizadeh, and
  Talwalkar]{mohri2018foundations}
Mehryar Mohri, Afshin Rostamizadeh, and Ameet Talwalkar.
\newblock \emph{Foundations of machine learning}.
\newblock MIT press, 2018.

\bibitem[Nilsback and Zisserman(2008)]{nilsback2008automated}
Maria-Elena Nilsback and Andrew Zisserman.
\newblock Automated flower classification over a large number of classes.
\newblock In \emph{2008 Sixth Indian conference on computer vision, graphics \&
  image processing}, pages 722--729. IEEE, 2008.

\bibitem[Oh et~al.(2021)Oh, Kim, and Yun]{oh2021fedbabu}
Jaehoon Oh, Sangmook Kim, and Se-Young Yun.
\newblock {FedBABU: Towards enhanced representation for federated image
  classification}.
\newblock \emph{arXiv preprint arXiv:2106.06042}, 2021.

\bibitem[Parkhi et~al.(2012)Parkhi, Vedaldi, Zisserman, and
  Jawahar]{parkhi2012cats}
Omkar~M Parkhi, Andrea Vedaldi, Andrew Zisserman, and CV Jawahar.
\newblock Cats and dogs.
\newblock In \emph{2012 IEEE conference on computer vision and pattern
  recognition}, pages 3498--3505. IEEE, 2012.

\bibitem[Paszke et~al.(2019)Paszke, Gross, Massa, Lerer, Bradbury, Chanan,
  Killeen, Lin, Gimelshein, Antiga, et~al.]{2019pytorch}
Adam Paszke, Sam Gross, Francisco Massa, Adam Lerer, James Bradbury, Gregory
  Chanan, Trevor Killeen, Zeming Lin, Natalia Gimelshein, Luca Antiga, et~al.
\newblock {Pytorch: An imperative style, high-performance deep learning
  library}.
\newblock \emph{Advances in neural information processing systems}, 32, 2019.

\bibitem[Peng et~al.(2019)Peng, Bai, Xia, Huang, Saenko, and
  Wang]{peng2019moment}
Xingchao Peng, Qinxun Bai, Xide Xia, Zijun Huang, Kate Saenko, and Bo Wang.
\newblock Moment matching for multi-source domain adaptation.
\newblock In \emph{Proceedings of the IEEE/CVF international conference on
  computer vision}, pages 1406--1415, 2019.

\bibitem[Phatak et~al.(2022)Phatak, Raghvendra, Tripathy, and
  Zhang]{phatak2022computing}
Abhijeet Phatak, Sharath Raghvendra, Chittaranjan Tripathy, and Kaiyi Zhang.
\newblock Computing all optimal partial transports.
\newblock In \emph{The Eleventh International Conference on Learning
  Representations}, 2022.

\bibitem[Radford et~al.(2021)Radford, Kim, Hallacy, Ramesh, Goh, Agarwal,
  Sastry, Askell, Mishkin, Clark, et~al.]{radford2021learning}
Alec Radford, Jong~Wook Kim, Chris Hallacy, Aditya Ramesh, Gabriel Goh,
  Sandhini Agarwal, Girish Sastry, Amanda Askell, Pamela Mishkin, Jack Clark,
  et~al.
\newblock {Learning transferable visual models from natural language
  supervision}.
\newblock In \emph{International conference on machine learning}, pages
  8748--8763. PMLR, 2021.

\bibitem[Sattler et~al.(2020)Sattler, M{\"u}ller, and
  Samek]{sattler2020clustered}
Felix Sattler, Klaus-Robert M{\"u}ller, and Wojciech Samek.
\newblock Clustered federated learning: Model-agnostic distributed multitask
  optimization under privacy constraints.
\newblock \emph{IEEE transactions on neural networks and learning systems},
  32\penalty0 (8):\penalty0 3710--3722, 2020.

\bibitem[Shamsian et~al.(2021)Shamsian, Navon, Fetaya, and
  Chechik]{shamsian2021personalized}
Aviv Shamsian, Aviv Navon, Ethan Fetaya, and Gal Chechik.
\newblock Personalized federated learning using hypernetworks.
\newblock In \emph{International Conference on Machine Learning}, pages
  9489--9502. PMLR, 2021.

\bibitem[Sinkhorn(1967)]{sinkhorn1967diagonal}
Richard Sinkhorn.
\newblock Diagonal equivalence to matrices with prescribed row and column sums.
\newblock \emph{The American Mathematical Monthly}, 74\penalty0 (4):\penalty0
  402--405, 1967.

\bibitem[Su et~al.(2022)Su, Yang, Li, and Xue]{su2022cross}
Shangchao Su, Mingzhao Yang, Bin Li, and Xiangyang Xue.
\newblock Cross-domain federated adaptive prompt tuning for clip.
\newblock \emph{arXiv preprint arXiv:2211.07864}, 2022.

\bibitem[Sun et~al.(2021)Sun, Huo, Yang, and Bai]{sun2021partialfed}
Benyuan Sun, Hongxing Huo, Yi Yang, and Bo Bai.
\newblock {Partialfed: Cross-domain personalized federated learning via partial
  initialization}.
\newblock \emph{Advances in Neural Information Processing Systems},
  34:\penalty0 23309--23320, 2021.

\bibitem[T~Dinh et~al.(2020)T~Dinh, Tran, and Nguyen]{t2020personalized}
Canh T~Dinh, Nguyen Tran, and Josh Nguyen.
\newblock Personalized federated learning with moreau envelopes.
\newblock \emph{Advances in Neural Information Processing Systems},
  33:\penalty0 21394--21405, 2020.

\bibitem[Tan et~al.(2022)Tan, Long, Ma, Liu, Zhou, and Jiang]{tan2022federated}
Yue Tan, Guodong Long, Jie Ma, Lu Liu, Tianyi Zhou, and Jing Jiang.
\newblock {Federated learning from pre-trained models: A contrastive learning
  approach}.
\newblock \emph{Advances in Neural Information Processing Systems},
  35:\penalty0 19332--19344, 2022.

\bibitem[Torous et~al.(2021)Torous, Gunsilius, and Rigollet]{torous2021optimal}
William Torous, Florian Gunsilius, and Philippe Rigollet.
\newblock An optimal transport approach to causal inference.
\newblock \emph{arXiv preprint arXiv:2108.05858}, 2021.

\bibitem[Tu et~al.(2022)Tu, Zhang, Kjellstr{\"o}m, and Zhang]{tu2022optimal}
Ruibo Tu, Kun Zhang, Hedvig Kjellstr{\"o}m, and Cheng Zhang.
\newblock Optimal transport for causal discovery.
\newblock \emph{arXiv preprint arXiv:2201.09366}, 2022.

\bibitem[Vahidian et~al.(2023)Vahidian, Morafah, Wang, Kungurtsev, Chen, Shah,
  and Lin]{vahidian2023efficient}
Saeed Vahidian, Mahdi Morafah, Weijia Wang, Vyacheslav Kungurtsev, Chen Chen,
  Mubarak Shah, and Bill Lin.
\newblock Efficient distribution similarity identification in clustered
  federated learning via principal angles between client data subspaces.
\newblock In \emph{Proceedings of the AAAI Conference on Artificial
  Intelligence}, pages 10043--10052, 2023.

\bibitem[Van~der Maaten and Hinton(2008)]{van2008visualizing}
Laurens Van~der Maaten and Geoffrey Hinton.
\newblock Visualizing data using t-sne.
\newblock \emph{Journal of machine learning research}, 9\penalty0 (11), 2008.

\bibitem[Wang et~al.(2019)Wang, Mathews, Kiddon, Eichner, Beaufays, and
  Ramage]{wang2019federated}
Kangkang Wang, Rajiv Mathews, Chlo{\'e} Kiddon, Hubert Eichner, Fran{\c{c}}oise
  Beaufays, and Daniel Ramage.
\newblock Federated evaluation of on-device personalization.
\newblock \emph{arXiv preprint arXiv:1910.10252}, 2019.

\bibitem[Xu et~al.(2023)Xu, Tong, and Huang]{xu2023personalized}
Jian Xu, Xinyi Tong, and Shao-Lun Huang.
\newblock Personalized federated learning with feature alignment and classifier
  collaboration.
\newblock \emph{arXiv preprint arXiv:2306.11867}, 2023.

\bibitem[Yang et~al.(2023)Yang, Wang, and Wang]{yang2023efficient}
Fu-En Yang, Chien-Yi Wang, and Yu-Chiang~Frank Wang.
\newblock Efficient model personalization in federated learning via
  client-specific prompt generation.
\newblock In \emph{Proceedings of the IEEE/CVF International Conference on
  Computer Vision}, pages 19159--19168, 2023.

\bibitem[Ye et~al.(2023)Ye, Xu, Wang, Xu, Chen, and Wang]{ye2023feddisco}
Rui Ye, Mingkai Xu, Jianyu Wang, Chenxin Xu, Siheng Chen, and Yanfeng Wang.
\newblock {FedDisco: Federated Learning with Discrepancy-Aware Collaboration}.
\newblock \emph{arXiv preprint arXiv:2305.19229}, 2023.

\bibitem[Zhao et~al.(2022)Zhao, Du, Li, Li, and Liu]{zhao2022reduce}
Haodong Zhao, Wei Du, Fangqi Li, Peixuan Li, and Gongshen Liu.
\newblock {Reduce communication costs and preserve privacy: Prompt tuning
  method in federated learning}.
\newblock \emph{arXiv preprint arXiv:2208.12268}, 2022.

\bibitem[Zhou et~al.(2022{\natexlab{a}})Zhou, Yang, Loy, and
  Liu]{zhou2022conditional}
Kaiyang Zhou, Jingkang Yang, Chen~Change Loy, and Ziwei Liu.
\newblock Conditional prompt learning for vision-language models.
\newblock In \emph{Proceedings of the IEEE/CVF Conference on Computer Vision
  and Pattern Recognition}, pages 16816--16825, 2022{\natexlab{a}}.

\bibitem[Zhou et~al.(2022{\natexlab{b}})Zhou, Yang, Loy, and
  Liu]{zhou2022learning}
Kaiyang Zhou, Jingkang Yang, Chen~Change Loy, and Ziwei Liu.
\newblock Learning to prompt for vision-language models.
\newblock \emph{International Journal of Computer Vision}, 130\penalty0
  (9):\penalty0 2337--2348, 2022{\natexlab{b}}.

\end{thebibliography}
}

\clearpage
\setcounter{page}{1}
\onecolumn
\begin{center}
{ \linespread{1.5} \selectfont
\textbf{\Large Global and Local Prompts Cooperation via Optimal Transport \\
for Federated Learning} \\
}
\Large Supplementary Materials
\end{center}
\appendix
\setcounter{table}{0}   
\setcounter{figure}{0}

\renewcommand{\thetable}{A\arabic{table}}
\renewcommand{\thefigure}{A\arabic{figure}}
\makeatletter

\startcontents[appendix]
\printcontents[appendix]{l}{1}{\section*{Supplementary organization:}\setcounter{tocdepth}{2}}

\section{Method Details}
\subsection{Efficient Scaling Dykstra’s Algorithm}\label{Dykstra’s_Algorithm}
As introduced in \cite{chang2023csot}, Problem (\ref{eq8}) can be solved by a fast implementation of Dykstra’s Algorithm by only performing matrix-vector multiplications, which is very similar to the widely-used and efficient Sinkhorn Algorithm \cite{cuturi2013sinkhorn}. The details of this algorithm are shown in Algorithm \ref{alg:Dykstras}.

\begin{algorithm}[h]
\caption{Efficient Scaling Dykstra's algorithm}
\label{alg:Dykstras}
\LinesNumbered
        \KwIn{Cost matrix $C$, marginal constraints vectors $\alpha$ and $\beta$, entropic regularization weight $\lambda$. }
        Initialize: $Q \gets e^{-C/\lambda}$, $v^{(0)} \gets \mathds{1}_{\beta}$, $\Delta_v = \infty$, $\epsilon=0.001$; \\
        Compute: $Q_\alpha \gets \frac{Q}{\text{diag}(\alpha)\mathds{1}_{|\alpha| \times |\beta|}}$,  $Q_\beta^\top \gets \frac{Q^\top}{\text{diag}(\beta)\mathds{1}_{|\beta| \times |\alpha|}}$; \\
        \For{$n = 1,2,3,\cdots$}
        {
            $u^{(n)} \gets \min \left( \frac{\mathds{1}_{|\alpha|}}{Q_\alpha v^{(n-1)}}, \mathds{1}_{|\alpha|} \right)$ ;\\
            $v^{(n)} \gets \frac{\mathds{1}_{|\beta|}}{Q_\beta^\top u^{(n)}}$;\\
            $\Delta_v = |v^{(n)} - v^{(n-1)}| $;\\
            \If{$\Delta_v < \epsilon$}
                {break}
            }
        \textbf{return} $\text{diag}(u^{(n)})Q\text{diag}(v^{(n)})$   
\end{algorithm}

\subsection{Training Process}\label{Training_Process}
Here, we provide detailed descriptions of the algorithm for our FedOTP, as shown in Algorithms \ref{alg:FedOTP}. For each communication round $t$, the selected clients perform local training by training global and local prompts $P_i^t = [P_g^t,p_{l,i}^t]$ through unbalanced OT at the same time. Then the updated global prompts $P_{g,i}^t$ are sent to the server for aggregation.

\begin{algorithm}[h]
\caption{FedOTP: Federated Prompts Cooperation via Optimal Transport}
\label{alg:FedOTP}
\LinesNumbered
    \KwIn{Communication rounds $T$, local epochs $R$, client number $N$, local dataset $D_i$, sample numbers $m_i$, pre-trained CLIP model $g(\cdot)$ and $h(\cdot)$, class number $K$, learning rate $\eta$, temperature of Softmax $\tau$. }
    Initialize parameters $P_i^0 = [P_g^0$, $P_{l,i}^0]$; \\
    \For{each communication round $t \in \{1,\cdots,T\}$}
        {
            Sample a client set $C^t \subset \{1,\cdots,N\}$ ;\\
        \For{each client $i \in C^t$}
        {
            Initialize $P_i^{t,0} = [P_g^{t-1},P_{l,i}^{t-1}]$;\\
            \For{each local epoch $r \in \{1,\cdots,R\}$}
            {
                Sample a mini-batch $B_i\in D_i$;\\
                Obtain a visual feature map $G_m$ with the visual encoder $g(x) (x\in B_i)$;\\
                Obtain textual features $H_k$ of each class with the textual encoder $\{h(P_{i,k}^{t,r-1})\}|_{k=1}^K$;\\
                Calculate the cost matrix $C_k = 1-G_m^\top H_k$ of each class;\\
                Solve Problem (\ref{eq8}) through Algorithm \ref{alg:Dykstras} and obtain Wasserstein distance $d_{C,k} = \left \langle T_k^\ast, C_k \right \rangle$;\\
                Calculate the classification probability $q(y=k|\textbf{x})=\frac{\exp((1-d_{C,k})/ \tau)}{\sum_{c=1}^{K}\exp((1-d_{C,c})/ \tau)}$;\\
                Update the parameters of prompts $P_i^{t,r} \gets P_i^{t,r-1} - \eta \nabla \mathcal{L}_{\mathcal{D}_i} (P_i^{t,r-1})$;\\
            }
        }
        Aggregate the global prompt $P_g^t = \sum_{i \in C_t}\frac{m_i}{\sum_{j \in C_t}m_j}P_{g,i}^{t,R}$;\\
    }
    \textbf{return} $P_i=[P_g, P_{l,i}]$  
\end{algorithm}

\begin{table*}[b]
\renewcommand\arraystretch{1.2}
\centering
\caption{The detailed statistics of datasets used in experiments.}
\label{Datasets_details}
\resizebox{0.9\textwidth}{!}{
\begin{tabular}{llllll}
\hline
\multicolumn{1}{c}{Dataset} & \multicolumn{1}{c}{Task}  & \multicolumn{1}{c}{Classes} & \multicolumn{1}{c}{Training Size} & \multicolumn{1}{c}{Testing Size} & \multicolumn{1}{c}{Domains}\\ \hline
Caltech101 \cite{fei2004learning}                     & Object recognition      & 100                   & 4,128                            & 2,465        &1 \\
Flowers102 \cite{nilsback2008automated}                    & Fine-grained flowers recognition      & 102                  & 4,093                            & 2,463      &1  \\
OxfordPets \cite{parkhi2012cats}                 & Fine-grained pets recognition & 37                         & 2,944                         & 3,669            &1   \\
Food101 \cite{bossard2014food}         & Fine-grained food recognition   &101    &50,500   &30,300        &1    \\
DTD \cite{cimpoi2014describing}                 & Texture recognition & 47                         & 2,820                       & 1,692              &1 \\ \hline
CIFAR-10 \cite{krizhevsky2009learning}                 & Image Classification & 10                        & 50,000                        & 10,000         &1      \\
CIFAR-100 \cite{krizhevsky2009learning}                & Image Classification & 100                         & 50,000                        & 10,000         &1      \\ \hline
DomainNet \cite{peng2019moment}                &  Image recognition    & 10                        & 18278                        & 4573              &6  \\ 
Office-Caltech10 \cite{gong2012geodesic}                 &  Image recognition   & 10                        & 2025                        & 508          &4     \\\hline
\end{tabular}
}
\end{table*}

\section{Experimental Details}
\label{Experimental_Details}

\subsection{Details of Dataset Setup}\label{Dataset_Setup}
\begin{figure}[htbp]
 \centering
	\begin{subfigure}{0.4\linewidth}
		\centering
		\includegraphics[width=0.8\linewidth]{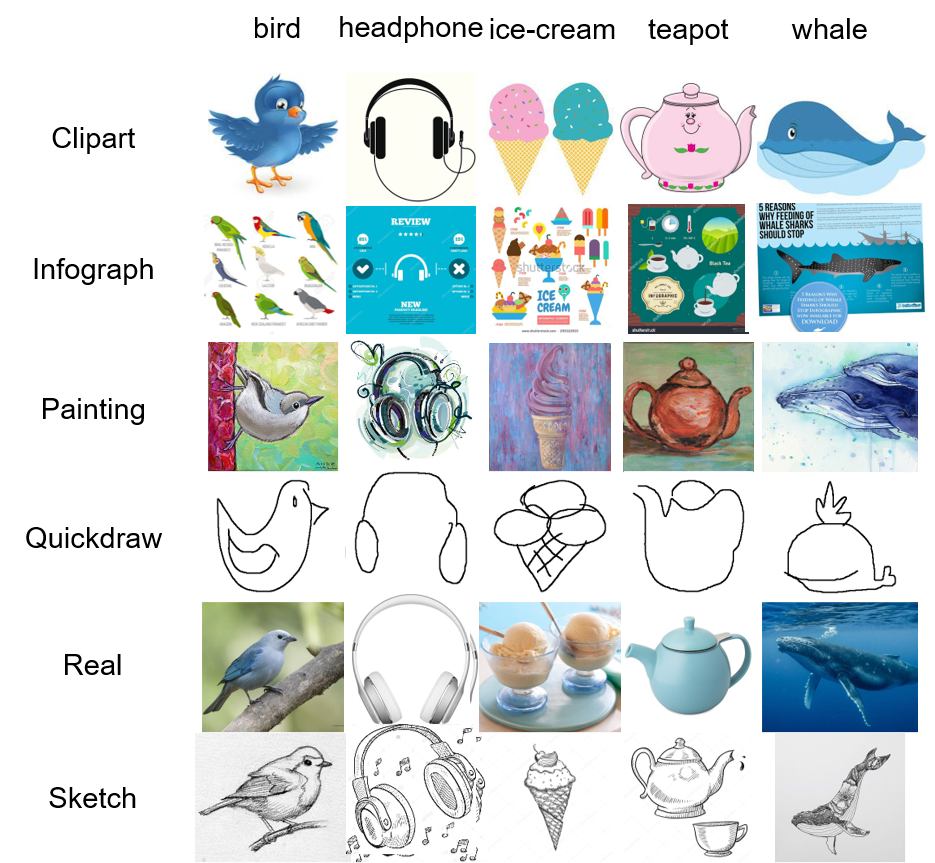}
		\caption{DomainNet}
	\end{subfigure}
 \centering
	\begin{subfigure}{0.59\linewidth}
		\centering
		\includegraphics[width=0.75\linewidth]{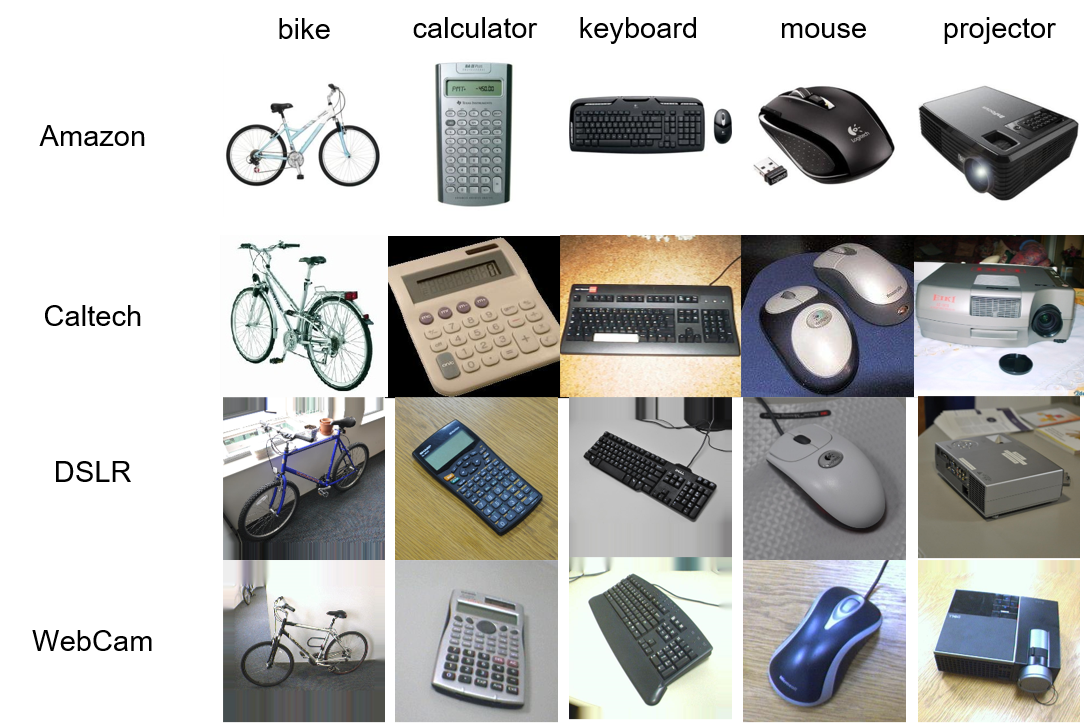}
		\caption{Office-Caltech10}
	\end{subfigure}
    \caption{Examples of raw instances from two datasets with multiple domains: DomainNet (left) and Office-Caltech10 (right). We present five classes for each dataset to show the feature shift across their sub-datasets.} 
    \label{example_feature_shift}  
\end{figure}

\begin{figure*}[ht]
\centering
\includegraphics[width=1\textwidth]{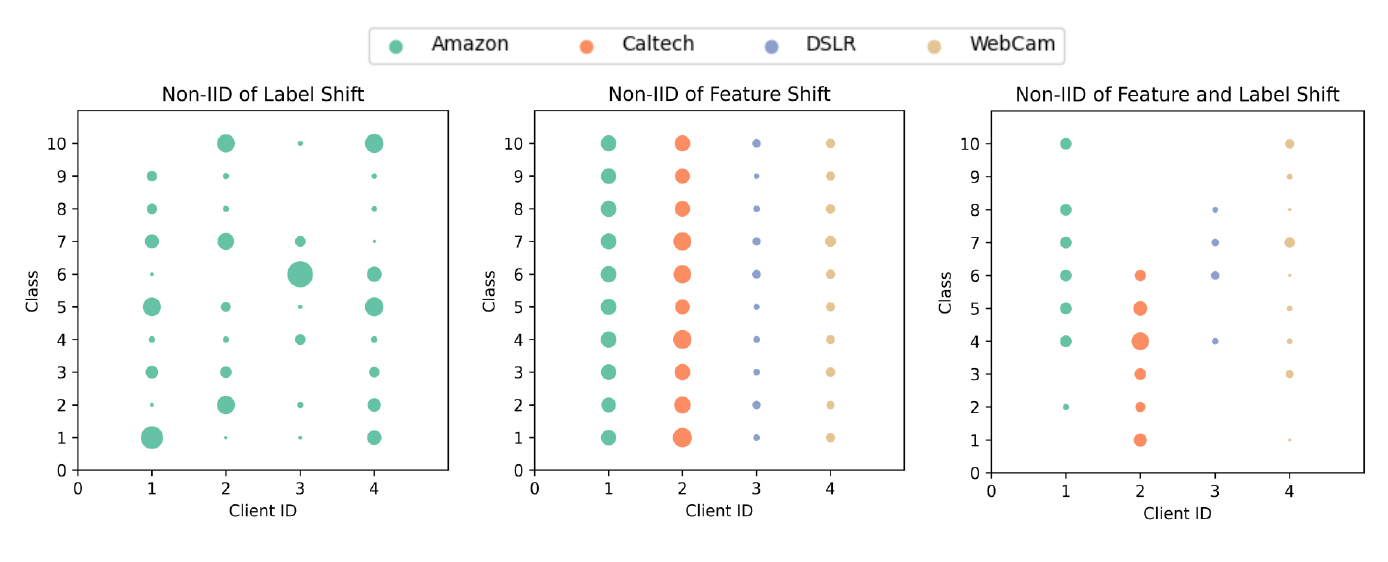} 
\caption{Visualization of three Non-IID settings on the Office-Caltech10 dataset. Each dot represents a set of samples within specific classes assigned to a client, with the dot size indicating the number of samples. The feature shifts are denoted by different colors.}
\label{noniid}
\end{figure*}

We select nine representative visual classification datasets as our benchmark. The detailed statistics of each dataset are shown in Table \ref{Datasets_details}, including the original tasks, the number of classes, the size of training and testing samples, and the number of domains.
As for datasets with multiple domains, Office-Caltech10 is a standard benchmark dataset consisting of four domains, namely Amazon, Caltech, DSLR, and WebCam, which are acquired using different camera devices or in different real environments with various backgrounds. DomainNet is a large-scale dataset consisting of six domains, namely Clipart, Infograph, Painting, Quickdraw, Real, and Sketch. We selected 10 classes from each of these two datasets for training. Some examples of raw instances of these two datasets can be found in Figure \ref{example_feature_shift}. For a clearer illustration, we visualize the three Non-IID settings employed in our paper in Figure \ref{noniid}.

\subsection{Implementation Details}\label{Implementation_Details}
All input images across datasets are resized to $224 \times 224$ pixels and further divided into $14 \times 14$ patches with a dimension of $768$. Regarding the hyperparameters for solving OT, we set the entropic regularization weight in Problem Eq. (\ref{eq8}) as $\lambda=0.1$ for all datasets. The maximum iteration number $n$ for Algorithm \ref{alg:Dykstras} is set to $100$, and we implement early stopping when the absolute update value $\Delta_v$ is less than $0.001$. For the setting of learnable prompts, the length of prompt vectors $s$ is set to $16$ with a dimension of $512$, ``end" token position, and ``random" initialization. Batch sizes are set to $32$ for training and $100$ for testing. All experiments are conducted with Pytorch \cite{2019pytorch} on NVIDIA A40 GPUs.

\section{Additional Experiments Results}
\subsection{Model Evaluation on Feature Shifts}\label{Model_Evaluation_on_Feature_Shifts}
In Table \ref{result_domain_datasets_1}, we compared the performance on Office-Caltech10 and DomainNet datasets under the presence of feature shift, where each client is assigned data from distinct domains while sharing the same label distribution. Our method achieved the highest average accuracies $99.16\%$ and $94.55\%$ on Office-Caltech10 and DomainNet, respectively.

\begin{table}[th]
\renewcommand\arraystretch{1.2}
\centering
\caption{Experimental results on Office-Caltech10 and DomainNet datasets with feature shift.}
\label{result_domain_datasets_1}
\resizebox{1\textwidth}{!}{
\begin{tabular}{lcccccccccccc} \hline
Datasets      & \multicolumn{5}{c}{Office-Caltech10}          & \multicolumn{7}{c}{DomainNet}     \\ \cline{2-13} 
Domains                                                    & A     & C     & D     & W     & Avg.                      & C                         & I                         & P                         & Q                         & R                         & S                         & Avg.                      \\ \hline
\multicolumn{13}{l}{\textbf{\textit{Local Training}}}      \\
Zero-Shot CLIP \cite{radford2021learning} & 19.3  & 18.2  & 21.9  & 18.6  & 19.50                     & 49.92                     & 47.15                     & 53.63                     & 31.3                      & 48.4                      & 50.18                     & 46.76                     \\
CoOp \cite{zhou2022learning}              & 96.38 & 97.24 & 100   & 98.31 & 97.98 & 98.32 & 83.01 & 98.18 & 82.37 & 98.21 & 97.70 & 92.95 \\ \hline
\multicolumn{13}{l}{\textbf{\textit{Prompt-based Federated Learning}}}      \\
PromptFL \cite{guo2023promptfl}           & 96.41 & 96.39 & 96.90 & 100   & 97.42                     & 98.23                     & 79.91                     & 97.89                     & 66.52                     & 96.83                     & 97.31                     & 89.45                     \\
PromptFL+FedProx \cite{li2020federated}   & 97.93 & 97.21 & 96.89 & 100   & 98.01                     & 98.45                     & 72.32                     & 96.00                     & 63.51                     & 96.08                     & 98.04                     & 87.40                     \\
FedOTP (Ours)                                              & \textbf{97.92} & \textbf{98.68} & \textbf{100}   & \textbf{100}   & \textbf{99.16}                     & \textbf{98.93}                     & \textbf{84.52}                     & \textbf{98.89}                     & \textbf{87.87}                     & \textbf{98.64}                     & \textbf{98.42}                     & \textbf{94.55}                     \\ \hline
\end{tabular}
}
\end{table}

\subsection{Model Evaluation on Feature $\&$ Label Shifts}\label{Feature_and_Label_Shifts}
In this set of experiments, we investigated scenarios involving both feature shifts and label shifts by dividing data within a domain into three clients based on the Dirichlet distribution with $\alpha=0.1$ for the Office-Caltech10 dataset. We calculated the mean and standard deviation of clients in the same domain, and the outcomes are presented in Table \ref{result_office}. Comparing these results with those in Table \ref{result_domain_datasets_1}, we can observe that the introduction of label shift leads to a performance decrease across all methods, with federated learning methods employing a shared prompt experiencing the most significant decline. In spite of this, our FedOTP consistently achieves the highest average accuracy, demonstrating its capability to utilize both global and local prompts to capture general domain-invariant and specific domain-specific knowledge for effective adaptation to extreme data heterogeneity.

\begin{table*}[h]
\renewcommand\arraystretch{1.2}
\centering
\caption{Experimental results on Office-Caltech10 dataset with feature $\&$ label shifts.}
\label{result_office}
\resizebox{0.85\textwidth}{!}{
\begin{tabular}{lccccc}
\hline
Datasets       & \multicolumn{5}{c}{Office-Caltech10}                                                                   \\ \cline{2-6} 
Domains         & Amazon           & Caltech           & DSLR           & Webcam           & Avg.        \\ \hline
\multicolumn{6}{l}{\textbf{\textit{Local Training}}}           \\
Zero-Shot CLIP \cite{radford2021learning}          & 8.45±1.49  & 6.01±4.25  & 12.92±9.15   & 6.48±4.82   & 8.46±6.26      \\
CoOp \cite{zhou2022learning} & \textbf{25.59±6.60} & \textbf{36.23±16.97}  & 30.30±5.18 & 22.56±5.46 & 28.67±11.13  \\ \hline
\multicolumn{6}{l}{\textbf{\textit{Prompt-based Federated Learning}}}           \\
PromptFL \cite{guo2023promptfl}      & 10.92±4.36   & 10.37±12.63  & 15.45±15.34  & 15.90±17.33  & 13.16±13.60  \\
PromptFL+FedProx \cite{li2020federated}    & 11.05±4.70 & 12.04±10.65 & 19.70±21.74   & 12.56±12.72  & 13.84±14.29    \\ 
FedOTP (Ours)           & 23.59±4.74 & 31.64±5.25  & \textbf{43.94±5.67}  & \textbf{35.51±9.19}  & \textbf{33.67±9.76}  \\ \hline
\end{tabular}
}
\end{table*}

\subsection{Effect of Parameter $\gamma$ in Unbalanced OT}\label{different_gamma}
In this subsection, we delved into the effect of parameter $\gamma$ in unbalanced OT, which regulates the mapping size of prompts on the feature map. We conducted experiments on the Pathological Non-IID setting across four datasets with varying numbers of shots and different values of the parameter $\gamma$ in our FedOTP. Specifically, we set $R=5$ and $T=10$ for these experiments. The results presented in Table \ref{result_gamma} reveal a notable trend: as the parameter $\gamma$ decreases, the overall performance initially increases and subsequently decreases. Interestingly, the majority of optimal results are observed at $\gamma=0.8$ or $\gamma=0.7$. This observation implies that the optimal alignment between global and local prompts and the feature map is achieved when the mapping size of prompts on the feature map is around $70\% - 80\%$. Consequently,  we adopt $\gamma=0.8$ in other experiments.

\begin{table*}[h]
\renewcommand\arraystretch{1.2}
\centering
\caption{Quantitative comparisons on the Pathological Non-IID setting across varying numbers of shots with different parameter $\gamma$ in our FedOTP over 10 clients.}
\label{result_gamma}
\resizebox{1\textwidth}{!}{
\begin{tabular}{llcccccc}
\hline
Dataset                     & shot number & 1                   & 0.9                 & 0.8                 & 0.7                 & 0.6                 & 0.5                 \\ \hline
\multirow{5}{*}{DTD}        & 1 shot      & 74.22±0.75          & 73.72±0.79          & 75.75±0.64          & 72.81±0.42          & \textbf{77.36±0.98} & 77.22±1.46          \\
                            & 2 shots     & 81.89±0.76          & 84.03±0.57          & 84.64±0.29          & \textbf{85.50±0.35} & 80.39±0.24          & 82.47±0.40          \\
                            & 4 shots     & 85.06±0.91          & 85.75±0.63          & 86.69±0.61          & \textbf{87.67±0.70} & 86.58±0.51          & 85.86±0.44          \\
                            & 8 shots     & 88.64±0.31          & 88.22±0.30          & 89.77±0.24          & \textbf{90.25±0.74} & 89.17±0.53          & 89.67±0.51          \\
                            & 16 shots    & 90.51±0.11          & 91.02±0.48          & \textbf{91.31±0.59} & 90.94±0.25          & 89.97±0.34          & 90.33±0.51          \\ \hline
\multirow{5}{*}{Caltech101} & 1 shot      & 89.68±1.19          & 92.13±0.58          & \textbf{92.54±0.71} & 91.53±0.55          & 90.10±1.74          & 90.67±0.44          \\
                            & 2 shots     & \textbf{95.05±0.49} & 93.89±0.35          & 94.45±0.32          & 94.37±0.43          & 93.89±0.65          & 94.68±0.92          \\
                            & 4 shots     & 96.02±0.36          & 96.64±0.41          & \textbf{97.02±0.36}          & 96.68±0.46 & 96.38±0.42          & 96.66±0.37          \\
                            & 8 shots     & 96.74±0.21          & 96.79±0.24          & 96.91±0.16          & 96.95±0.26          & 97.22±0.33          & \textbf{97.34±0.18} \\
                            & 16 shots    & 97.72±0.14          & 97.69±0.17          & 97.39±0.11          & 97.58±0.23          & 97.74±0.19          & \textbf{97.83±0.18} \\ \hline
\multirow{5}{*}{Flowers102} & 1 shot      & 86.68±1.93          & 85.77±0.74          & 87.42±0.92          & 88.43±0.90          & \textbf{89.14±1.18} & 85.56±1.21          \\
                            & 2 shots     & 93.09±1.26          & \textbf{93.96±0.48} & 93.31±0.55          & 93.13±0.26          & 93.70±0.49          & 93.56±0.86          \\
                            & 4 shots     & 95.46±0.55          & \textbf{96.23±0.44} & 95.51±0.30          & 95.89±0.50          & 96.17±0.47          & 96.16±0.34          \\
                            & 8 shots     & 97.53±0.24          & 97.49±0.19          & \textbf{98.23±0.32} & 98.11±0.27          & 97.24±0.28          & 97.40±0.64          \\
                            & 16 shots    & 98.86±0.15          & 98.30±0.55          & \textbf{99.11±0.11} & 98.88±0.17          & 99.03±0.12          & 98.93±0.18          \\ \hline
\multirow{5}{*}{OxfordPets} & 1 shot      & 95.82±1.16          & 94.26±0.38          & 96.18±0.71          & \textbf{96.37±0.79} & 94.21±0.53          & 95.97±0.42          \\
                            & 2 shots     & \textbf{97.73±0.57} & 96.12±0.32          & 97.50±1.02          & 97.49±0.45          & 97.60±0.40          & 97.27±0.19          \\
                            & 4 shots     & 98.11±1.15          & 98.46±0.64          & \textbf{98.82±0.11}          & 98.51±0.10  & 98.52±0.25          & 98.43±0.28          \\
                            & 8 shots     & 98.73±0.27          & \textbf{99.02±0.38} & 98.71±0.16          & 98.74±0.18          & 98.54±0.22          & 98.63±0.13          \\
                            & 16 shots    & 99.04±0.16          & 98.82±0.25          & \textbf{99.27±0.23} & 99.21±0.19          & 99.04±0.27          & 98.81±0.21          \\ \hline
\end{tabular}
}
\end{table*}

\subsection{Effect of Heterogeneity in Label Distribution}\label{different_alpha}
In addressing the core challenge of data heterogeneity in personalized federated learning, FedOTP consistently outperforms benchmark methods across various settings. Now, we investigated the effect of heterogeneity in label distribution by considering a range of $\alpha$ values of Dirichlet distribution, specifically $\alpha \in \{0.1, 0.3, 0.5, 1, 5, 10\}$ for CIFAR-100 datasets. It's worth noting that a smaller $\alpha$ implies a higher degree of data heterogeneity in these experiments. The results presented in Table \ref{result_cifar100} clearly indicate that as the degree of data heterogeneity increases, the performance of federated learning methods with a shared prompt decreases while the performance of CoOp and our FedOTP improves. Among these methods, FedOTP outperforms them in every case and demonstrates remarkable robustness. These findings underscore the effectiveness of FedOTP in overcoming label distribution heterogeneity across a diverse range of scenarios.

\begin{table*}[h]
\renewcommand\arraystretch{1.2}
\centering
\caption{Quantitative comparisons on CIFAR-100 dataset with different $\alpha$ of the Dirichlet setting.}
\label{result_cifar100}
\resizebox{1\textwidth}{!}{
\begin{tabular}{lcccccc}
\hline
Dataset                       & \multicolumn{6}{c}{CIFAR-100}                                               \\ \cline{2-7} 
$\# \alpha$                          & 0.1        & 0.3        & 0.5        & 1          & 5          & 10         \\ \hline
\multicolumn{7}{l}{\textbf{\textit{Local Training}}}                                                                          \\
Zero-Shot CLIP \cite{radford2021learning}               & 65.22±0.32 & 64.92±0.53 & 65.78±0.41 & 63.93±0.16 & 64.01±0.27 & 65.07±0.35 \\
CoOp \cite{zhou2022learning}                         & 62.01±0.29 & 74.83±0.45 & 51.72±0.42 & 47.03±0.37 & 41.03±0.23 & 41.37±0.19 \\ \hline
\multicolumn{7}{l}{\textbf{\textit{Prompt-based Federated Learning}}}                                                         \\
PromptFL \cite{guo2023promptfl}                     & 72.45±0.64 & 73.67±0.56 & 74.37±0.18 & 73.95±0.14 & 74.68±0.05 & 74.43±0.08 \\
PromptFL+FedProx \cite{li2020federated}             & 72.57±0.54 & 71.11±0.91 & 74.45±0.19 & 74.19±0.06 & 74.23±0.09 & 74.53±0.07 \\
FedOTP (Similarity Averaging) & 78.68±0.17 & 75.70±0.27 & 75.28±0.12 & 74.88±0.16 & 74.48±0.05 & 74.31±0.39 \\
FedOTP (Classical OT)         & 79.93±0.19 & 77.86±0.09 & 75.76±0.12 & 75.38±0.08 & 75.01±0.05 & 74.73±0.05 \\
FedOTP (Unbalanced OT)        & \textbf{80.56±0.12} & \textbf{78.03±0.08} & \textbf{76.75±0.10} & \textbf{76.17±0.13} & \textbf{75.75±0.03} & \textbf{75.52±0.06} \\ \hline
\end{tabular}
}
\end{table*}

\subsection{Learning Curves}
\begin{figure*}[ht]
\centering
\includegraphics[width=1\textwidth]{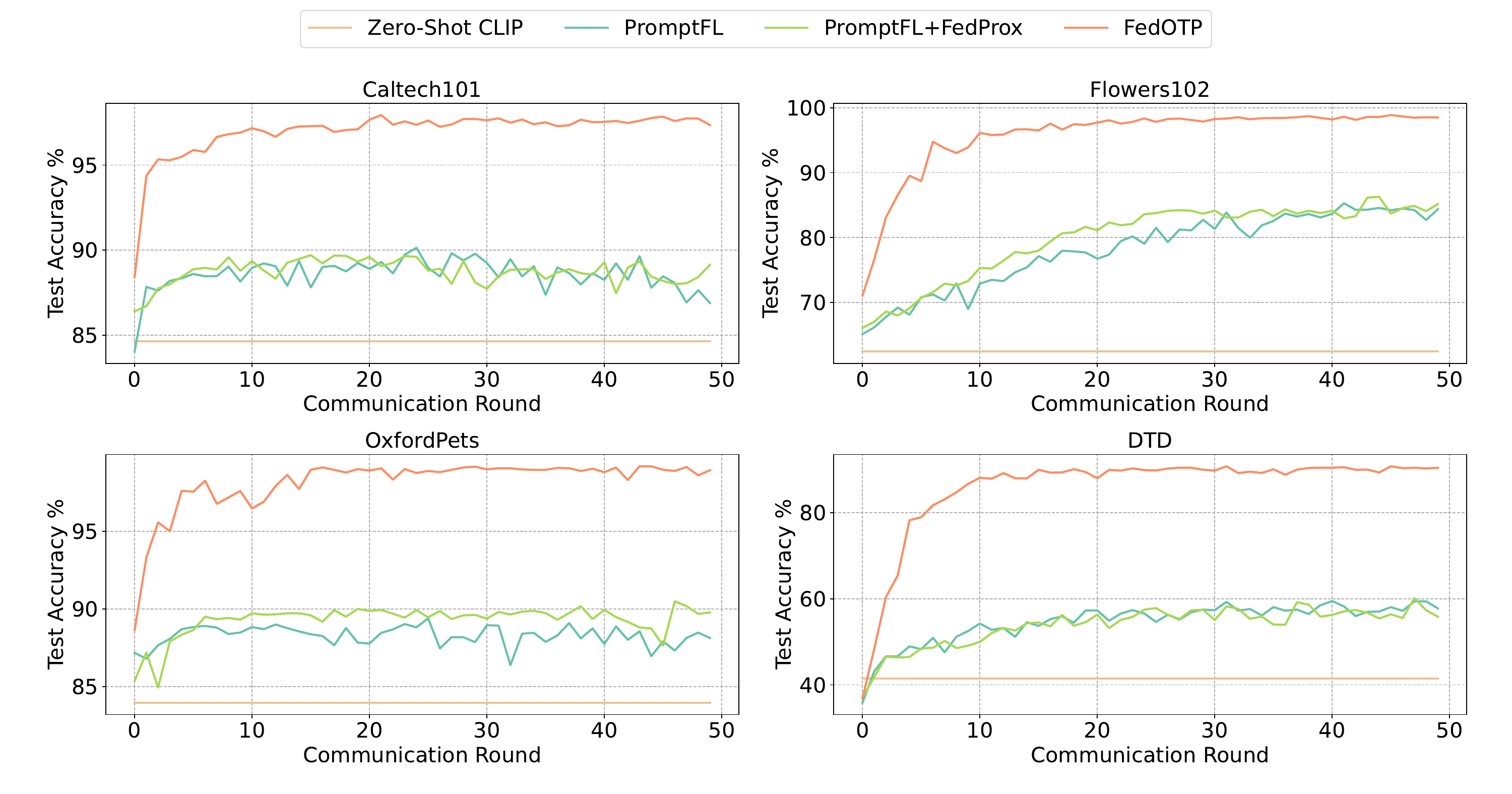} 
\caption{Accuracy curves and convergence behavior of FedOTP and other baselines on four datasets over 10 clients.}
\label{learning_curve}
\end{figure*}

To assess the convergence of our method, we plotted test accuracy curves with $R=1$ and $T=50$ for different methods across four datasets, as illustrated in 
Figure \ref{learning_curve}. Compared to other methods, FedOTP exhibits notable characteristics of accelerated convergence and enhanced stability, evident from the smaller fluctuations in test accuracy.

\section{Visualization}
\subsection{Visualizations of Transport Plans}\label{Visualizations_of_Transport_Plans}
\begin{figure}[htbp]
 \centering
	\begin{subfigure}{1\linewidth}
		\centering
		\includegraphics[width=1\linewidth]{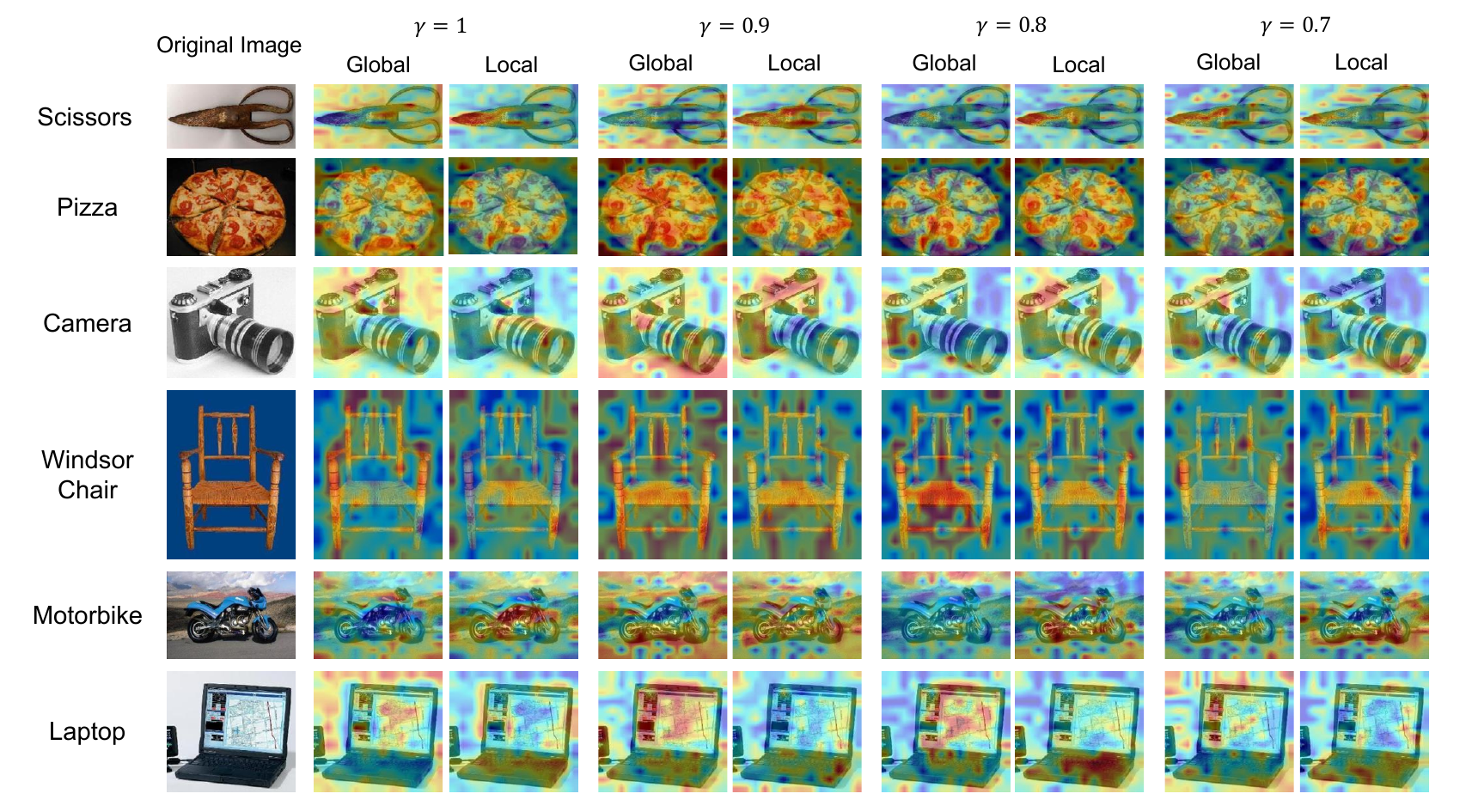}
		\caption{Artifacts in Caltech101 dataset.}
	\end{subfigure}
 \centering
	\begin{subfigure}{1\linewidth}
		\centering
		\includegraphics[width=1\linewidth]{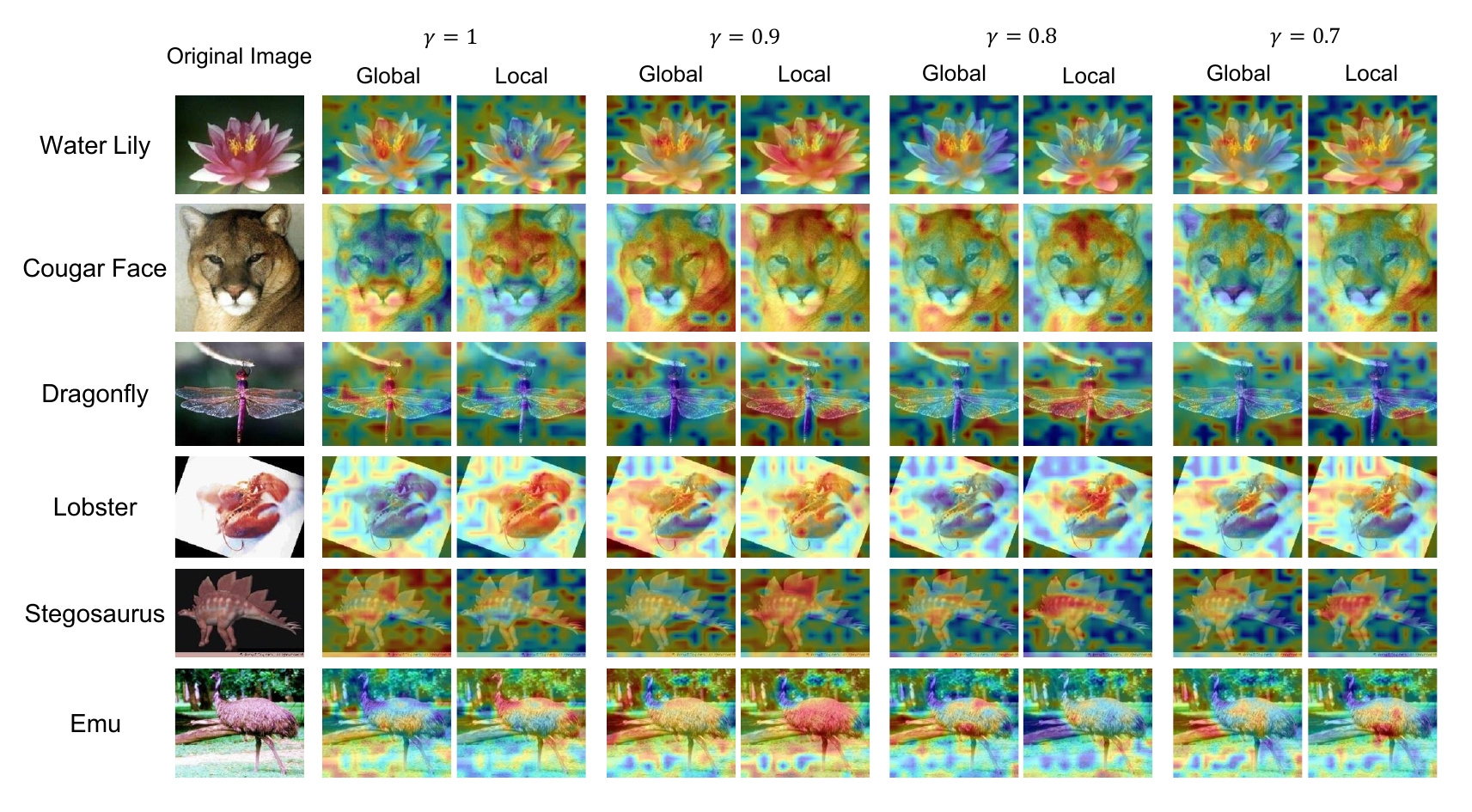}
		\caption{Organisms in Caltech101 dataset.}
	\end{subfigure}
    \caption{Heatmaps of transport plans related to global and local prompts of FedOTP with different $\gamma$ in Caltech101 dataset. ``Global" denotes the transport plans related to global prompts and ``Local" refers to local prompts.} 
    \label{heatmap1}  
\end{figure}

To facilitate a comparative analysis between FedOTP, PromptFL, and Local models, we examined visualizations of similarity between textual features and feature maps in Figure \ref{heatmap}. Here, we provided visualization examples showcasing transport plans $T$ associated with global and local prompts of FedOTP across different $\gamma$ values. We converted each transport plan into colorful heatmaps, resized and overlayed them on the original image. The comparisons between heatmaps of transport plans with different $\gamma$ in Caltech101 dataset is presented in Figure \ref{heatmap1}. Upon observation, we noted that when $\gamma=1$, the transport plans related to global and local prompts exhibit complementary, as each image patch is assigned to prompts due to the equality constraints of OT. This may result in the integration of objects and backgrounds on the global part, observable in classes like ``Camera" and ``Laptop". In contrast, as $\gamma$ decreases, prompts focus on a smaller range of patches and primarily center on the patches of main objects rather than backgrounds, further supporting the claim that FedOTP can effectively regulate the mapping size of prompts on the feature map.

usiv
\subsection{T-SNE Projection of Prompts.}
\begin{figure}[ht]
\centering
\includegraphics[width=0.5\textwidth]{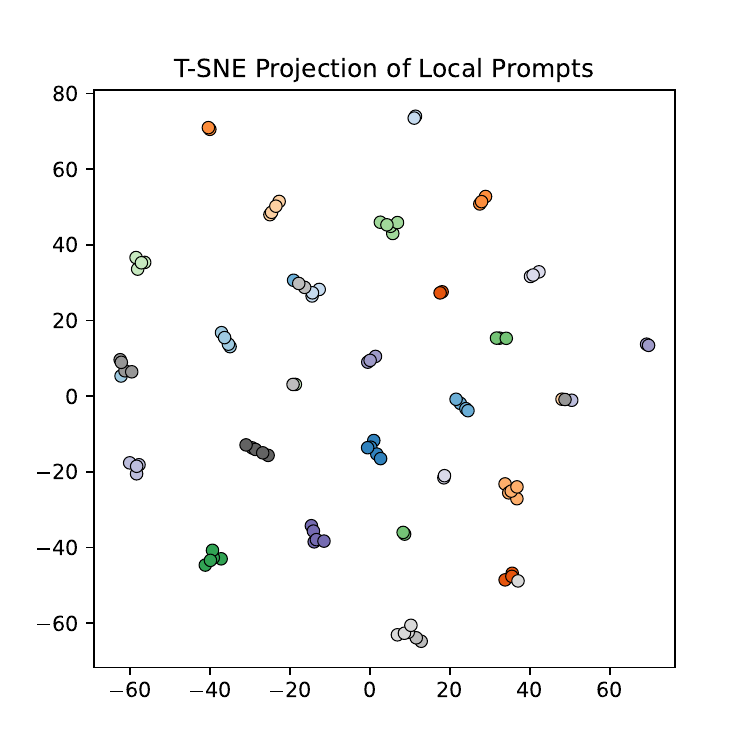} 
\caption{T-SNE projection of local prompts from FedOTP in CIFAR-100 dataset.}
\label{tsne}
\end{figure}

To examine how the learned prompts form a meaningful representation across the client space, we employed the t-SNE algorithm \cite{van2008visualizing} to project prompts onto a $2D$ plane. Following \cite{shamsian2021personalized, li2023fedtp}, we divided CIFAR-100 dataset into $100$ clients. In detail, each coarse label was assigned to five clients, and the corresponding fine labels were uniformly distributed among those selected clients. 
After training these clients with FedOTP, we visualized their local prompts obtained from local training, and we used different colors to represent various coarse labels. As shown in Figure \ref{tsne}, local prompts from clients with the same coarse label are clustered together and positioned far from those with different coarse labels. These results further illustrate that in FedOTP, the acquired local prompts are tailored to capture client-specific category characteristics.

\section{Generalization Bound}
\label{sec:Generalization_Bound}

\subsection{Key Lemmas}
\newtheorem{lemma}{Lemma}

\begin{lemma}[McDiarmid's Inequality \cite{mohri2018foundations}] \label{lemma1}
Let $X_1,\cdots,X_n$ be independent random variables, where $X_i$ has range $\mathcal{X}_i$. Let $g:\mathcal{X}_1 \times \cdots \times \mathcal{X}_n\rightarrow \mathbb{R}$ be any function with the $(a_1,\cdots,a_n)-$bounded difference property: for every $i=1,\cdots,n$ and $x_1,\cdots,x_n,x_i^{\prime} \in \mathcal{X}_1 \times \cdots \times \mathcal{X}_n$, we have
\begin{equation}
\centering
 \sup \limits_{x_i \in \mathcal{X}_i} |g(x_1,\cdots,x_i,\cdots,x_n)-g(x_1,\cdots,x_i^{\prime},\cdots,x_n)| \leq a_i.
\end{equation}
Then for any $\varepsilon >0$, 
\begin{equation}
\centering
\begin{aligned}
 \mathbb{P}[g(X_1,\cdots,X_n)-\mathbb{E}[g(X_1,\cdots,X_n)]\geq \varepsilon] \leq \exp\left(-\frac{2\varepsilon^2}{\sum_{i=1}^{n}a_i^2}\right).
\end{aligned}
\end{equation}
\end{lemma}

\begin{lemma}[Rademacher Complexity \cite{mohri2018foundations}] \label{lemma2}
Given a space $\mathcal{B}$ and a fixed distribution $D_B$, let $\{b_1,\cdots,b_m\}$ be a set of examples drawn i.i.d. from $D_B$. Let $\mathcal{F}$ be a class of functions $f: B \rightarrow \mathbb{R}$, and the Rademacher Complexity of $\mathcal{F}$ is defined as follows:
\begin{equation}
    \centering
    \Re_{D_B}(\mathcal{B}) = \frac{1}{m} \mathbb{E}_{\sigma} \left[\sup\limits_{b\in B}\sum_{i=1}^{m}\sigma_i b_i \right].
\end{equation}
where $\sigma_1,\cdots,\sigma_m$ are independent random variables uniformly chosen from $\{-1,1\}$.
\end{lemma}

\subsection{Proof of Theorem 1}
\newenvironment{proof}{{\it Proof:\quad}}{\hfill $\blacksquare$\par}
\begin{proof}
For the left side of Theorem \ref{theorem1}, we have
\begin{equation}\label{eq16}
    \begin{aligned}
    \centering
    &\left|\sum_{i=1}^N \frac{m_i}{M}\left(\mathcal{L}_{\hat{\mathcal{D}}_i}(\hat{P}_g,\hat{P}_{l,i})-\mathcal{L}_{\mathcal{D}_i}(P_g^*,P_{l,i}^*) \right)\right| \\
    &=\bigg|\sum_{i=1}^N \frac{m_i}{M}\Big(\mathcal{L}_{\hat{\mathcal{D}}_i}(\hat{P}_g,\hat{P}_{l,i}) -\mathcal{L}_{\hat{\mathcal{D}}_i}(P_g^*,P_{l,i}^*) + \mathcal{L}_{\hat{\mathcal{D}}_i}(P_g^*,P_{l,i}^*) -\mathcal{L}_{\mathcal{D}_i}(P_g^*,P_{l,i}^*) \Big) \bigg| \\
    &\leq \left|\sum_{i=1}^N \frac{m_i}{M}\left(\mathcal{L}_{\hat{\mathcal{D}}_i}(\hat{P}_g,\hat{P}_{l,i})-\mathcal{L}_{\hat{\mathcal{D}}_i}(P_g^*,P_{l,i}^*) \right)\right| + \left|\sum_{i=1}^N \frac{m_i}{M}\left(\mathcal{L}_{\hat{\mathcal{D}}_i}(P_g^*,P_{l,i}^*)-\mathcal{L}_{\mathcal{D}_i}(P_g^*,P_{l,i}^*) \right)\right| .
    \end{aligned}
\end{equation}

The objective function is partitioned into two components, and we will bound each of them independently. Concerning the first part in Eq. (\ref{eq16}), assuming Assumptions \ref{assum1} and \ref{assum2} hold, we obtain
\begin{equation}\label{eq17}
    \begin{aligned}
    \centering
    &\left|\sum_{i=1}^N \frac{m_i}{M} \left(\mathcal{L}_{\hat{\mathcal{D}}_i}(\hat{P}_g,\hat{P}_{l,i})-\mathcal{L}_{\hat{\mathcal{D}}_i}(P_g^*,P_{l,i}^*) \right)\right| \\
    &\leq \bigg|\sum_{i=1}^N \frac{m_i}{M} \Big(\mathcal{L}_{\hat{\mathcal{D}}_i}(\hat{P}_g,\hat{P}_{l,i})-\mathcal{L}_{\hat{\mathcal{D}}_i}(\hat{P}_g,P_{l,i}^*) + \mathcal{L}_{\hat{\mathcal{D}}_i}(\hat{P}_g,P_{l,i}^*) - \mathcal{L}_{\hat{\mathcal{D}}_i}(P_g^*,P_{l,i}^*) \Big)\bigg| \\
    &\leq \bigg|\sum_{i=1}^N \frac{m_i}{M} \Big(\mathcal{L}_{\hat{\mathcal{D}}_i}(\hat{P}_g,\hat{P}_{l,i})-\mathcal{L}_{\hat{\mathcal{D}}_i}(\hat{P}_g,P_{l,i}^*) \Big)\bigg| + \bigg|\sum_{i=1}^N \frac{m_i}{M} \Big(\mathcal{L}_{\hat{\mathcal{D}}_i}(\hat{P}_g,P_{l,i}^*) - \mathcal{L}_{\hat{\mathcal{D}}_i}(P_g^*,P_{l,i}^*) \Big)\bigg| \\
    &\leq \sum_{i=1}^N \frac{m_i}{M} \mathbb{E}_{(x_i^{j}, y_i^{j}) \in {D}_i} \bigg| \ell (f(\hat{P}_g,\hat{P}_{l,i}; x_i^{j}), y_i^{j}) - \ell (f(\hat{P}_g,P_{l,i}^\ast; x_i^{j}), y_i^{j}) \bigg| \\
    &+ \sum_{i=1}^N \frac{m_i}{M} \mathbb{E}_{(x_i^{j}, y_i^{j}) \in {D}_i} \bigg| \ell (f(\hat{P}_g,P_{l,i}^\ast; x_i^{j}), y_i^{j}) - \ell (f(P_g^\ast,P_{l,i}^\ast; x_i^{j}), y_i^{j}) \bigg| \\
    &\leq \sum_{i=1}^N \frac{m_i}{M} L \Big( \| f(\hat{P}_g,\hat{P}_{l,i}) - f(\hat{P}_g,P_{l,i}^\ast) \| + \| f(\hat{P}_g,P_{l,i}^\ast) - f(P_g^\ast,P_{l,i}^\ast) \| \Big) \\
    &\leq \sum_{i=1}^N \frac{m_i}{M} \Big(L L_g \|\hat{P}_{l,i}-P_{l,i}^\ast\| + L L_{l,i} \|\hat{P}_g-P_g^\ast\| \Big) \\
    &\leq L L_g A_g + L \sum_{i=1}^N \frac{m_i}{M} L_{l,i} A_{l,i} 
    \leq L L_g A_g + L \Big(\sum_{i=1}^N \frac{m_i}{M} \Big)\Big(\sum_{i=1}^N  L_{l,i} A_{l,i}\Big) \\
    &\leq  L L_g A_g + L \sqrt{\Big(\sum_{i=1}^N  L^2_{l,i} \Big) \Big(\sum_{i=1}^N  A^2_{l,i}\Big)}.
    \end{aligned}
\end{equation}

For the second part in Eq. (\ref{eq16}), replacing $g(\cdot)$ with $\sum_{i=1}^N \frac{m_i}{M}\left(\mathcal{L}_{\hat{\mathcal{D}}_i}(P_g^*,P_{l,i}^*)-\mathcal{L}_{\mathcal{D}_i}(P_g^*,P_{l,i}^*) \right)$ in Lemma \ref{lemma1}, and setting $\delta=\exp\left(-2\varepsilon^2/\sum_{i=1}^n a_i^2\right)$, with a probability at least $1-\delta$, the following inequality holds,
\begin{equation}\label{eq18}
    \begin{aligned}
    \centering
    &\left|\sum_{i=1}^N \frac{m_i}{M}\left(\mathcal{L}_{\hat{\mathcal{D}}_i}(P_g^*,P_{l,i}^*)-\mathcal{L}_{\mathcal{D}_i}(P_g^*,P_{l,i}^*) \right)\right| \\
    &\leq \mathbb{E} \left[ \sum_{i=1}^N \frac{m_i}{M}\left(\mathcal{L}_{\hat{\mathcal{D}}_i}(P_g^*,P_{l,i}^*)-\mathcal{L}_{\mathcal{D}_i}(P_g^*,P_{l,i}^*) \right) \right] + \sqrt{\frac{M}{2}log\frac{N}{\delta}}.
    \end{aligned}
\end{equation}

Utilizing Lemma \ref{lemma2} and the results in \cite{mansour2020three}, we can get
\begin{equation}\label{eq19}
    \begin{aligned}
    \centering
    &\mathbb{E} \left[ \sum_{i=1}^N \frac{m_i}{M}\left(\mathcal{L}_{\hat{\mathcal{D}}_i}(P_g^*,P_{l,i}^*)-\mathcal{L}_{\mathcal{D}_i}(P_g^*,P_{l,i}^*) \right) \right] \leq \sum_{i=1}^N\frac{m_i}{M} \Re_{\mathcal{D}_i}(\mathcal{H}) \\
    &\leq \sum_{i=1}^N\frac{m_i}{M} \sqrt{\frac{dN}{m_i}log\frac{em_i}{d}} \leq \sum_{i=1}^N\frac{m_i}{M}\sqrt{\frac{dN}{m_i}log\frac{eM}{d}} \leq \sqrt{\frac{dN}{M}log\frac{eM}{d}}.
    \end{aligned}
\end{equation}

Combining the results in Eq. (\ref{eq18}) and Eq. (\ref{eq19}), we can get
\begin{equation}\label{eq20}
    \begin{aligned}
    \centering
    \left|\sum_{i=1}^N \frac{m_i}{M}\left(\mathcal{L}_{\hat{\mathcal{D}}_i}(P_g^*,P_{l,i}^*)-\mathcal{L}_{\mathcal{D}_i}(P_g^*,P_{l,i}^*) \right)\right| \leq \sqrt{\frac{M}{2}log\frac{N}{\delta}} + \sqrt{\frac{dN}{M}log\frac{eM}{d}}.
    \end{aligned}
\end{equation}

Summarizing the results above, we can obtain 
\begin{equation}\label{eq21}
    \begin{aligned}
    \centering
     \left|\sum_{i=1}^N \frac{m_i}{M}\left(\mathcal{L}_{\hat{\mathcal{D}}_i}(\hat{P}_g,\hat{P}_{l,i})-\mathcal{L}_{\mathcal{D}_i}(P_g^*,P_{l,i}^*) \right)\right| 
     &\leq \sqrt{\frac{M}{2}log\frac{N}{\delta}} + \sqrt{\frac{dN}{M}log\frac{eM}{d}} + L L_g A_g + L \sqrt{\Big(\sum_{i=1}^N  L^2_{l,i} \Big) \Big(\sum_{i=1}^N  A^2_{l,i}\Big)} \\
     &= \sqrt{\frac{M}{2}log\frac{N}{\delta}} + \sqrt{\frac{dN}{M}log\frac{eM}{d}} + L( L_g A_g + L_lA_l),
    \end{aligned}
\end{equation}
where we denote $L_l=\sqrt{\sum_{i=1}^N  L^2_{l,i}}$ and $A_l=\sqrt{\sum_{i=1}^N  A^2_{l,i}}$ for simplicity.
\end{proof}

\end{document}